\documentclass[11pt]{article}

\usepackage[preprint]{acl}

\usepackage{times}
\usepackage{latexsym}
\usepackage{listings}
\usepackage[T1]{fontenc}

\usepackage[utf8]{inputenc}

\usepackage{microtype}

\usepackage{inconsolata}

\usepackage{graphicx}

\usepackage{amsmath} 
\usepackage{amssymb}
\usepackage{booktabs}
\usepackage{multirow}
\usepackage{array}
\usepackage[most]{tcolorbox}
\usepackage{multicol}
\usepackage{makecell}
\usepackage{utfsym}

\usepackage{algorithm,algorithmic}

\usepackage[T1]{fontenc}
\usepackage[utf8]{inputenc}
\usepackage{times}
\usepackage{microtype}

\title{Adaptive LLM-Symbolic Reasoning via Dynamic Logical Solver Composition}

\author{
    Lei Xu\textsuperscript{\rm 1,\rm 2}\thanks{Equal contribution.}, 
    Pierre Beckmann\textsuperscript{\rm 1,\rm 2}\footnotemark[1],
    Marco Valentino\textsuperscript{\rm 3},
    André Freitas\textsuperscript{\rm 1,\rm 4,\rm 5}
\\
    \textsuperscript{\rm 1}Idiap Research Institute, Switzerland\\
    \textsuperscript{\rm 2}École Polytechnique Fédérale de Lausanne (EPFL), Switzerland\\
    \textsuperscript{\rm 3}School of Computer Science, University of Sheffield\\
    \textsuperscript{\rm 4}Department of Computer Science, University of Manchester, United Kingdom\\
    \textsuperscript{\rm 5}Cancer Biomarker Centre, CRUK Manchester Institute, United Kingdom
\\
 {
   \texttt{\{firstname.lastname\}@idiap.ch}
 }
 \\
}

\newcommand{\Router}{\texttt{Router}}

\begin{document}
\maketitle
\begin{abstract}
Neuro-symbolic NLP methods aim to leverage the complementary strengths of large language models and formal logical solvers. However, current approaches are mostly static in nature, i.e., the integration of a target solver is predetermined at design time, hindering the ability to employ diverse formal inference strategies. To address this, we introduce an adaptive, multi-paradigm, neuro-symbolic inference framework that: (1) automatically identifies formal reasoning strategies from problems expressed in natural language; and (2) dynamically selects and applies specialized formal logical solvers via autoformalization interfaces. Extensive experiments on individual and multi-paradigm reasoning tasks support the following conclusions: LLMs are effective at predicting the necessary formal reasoning strategies with an accuracy above 90\%. This enables flexible integration with formal logical solvers, resulting in our framework outperforming competing baselines—by 27\% and 6\% compared to GPT-4o and DeepSeek-V3.1, respectively. Moreover, adaptive reasoning can even positively impact pure LLM methods, yielding gains of 10\%, 5\%, and 6\% on zero-shot, CoT, and CoT$_{sym}$ settings with GPT-4o. Finally, although smaller models struggle with adaptive neuro-symbolic reasoning, post-training offers a viable path to improvement. Overall, this work establishes the foundations for adaptive LLM-symbolic reasoning, offering a path forward for unifying material and formal inferences on heterogeneous reasoning challenges.
\end{abstract}

\section{Introduction}

\begin{figure*}
    \centering
    \includegraphics[width=\linewidth]{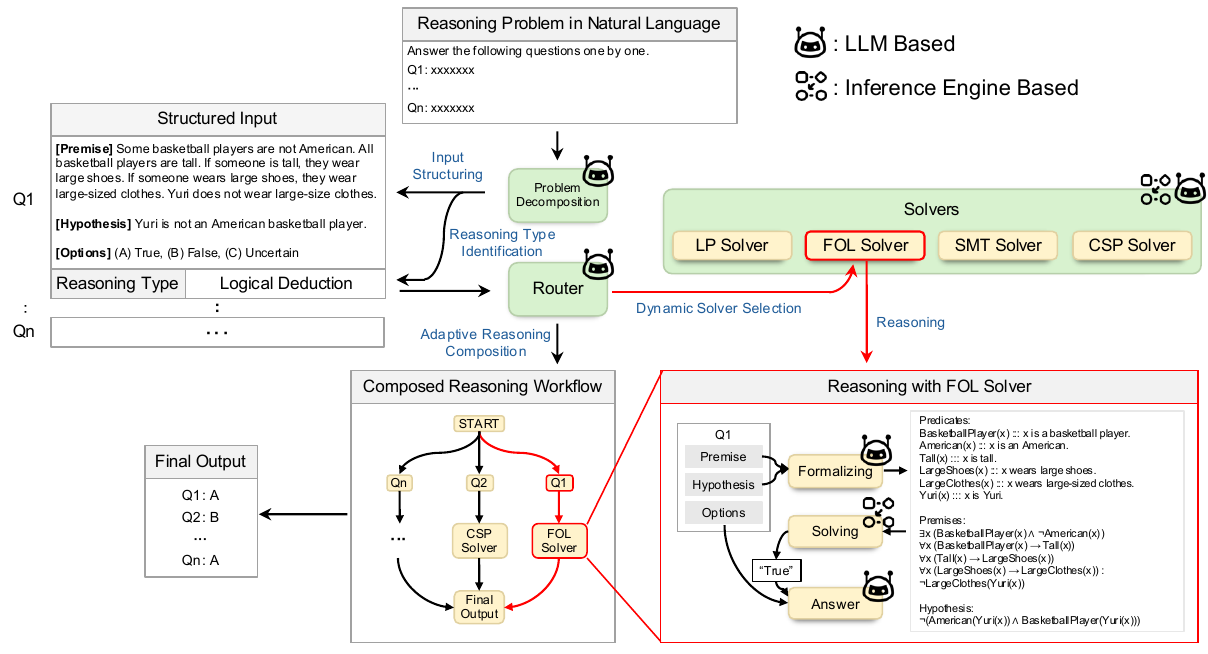}
    \caption{Overview of the proposed adaptive symbolic reasoning framework. Given a natural language reasoning problem, the system first performs problem decomposition to extract structured components and identify the corresponding reasoning type (LP, FOL, CSP, SMT). Based on this analysis, a Router dynamically selects the appropriate solver for each problem and orchestrates the reasoning process. Each solver performs autoformalization on the structured input and conducts formal reasoning to produce verified answers. Unlike static approaches that predetermine solver integration, our framework adaptively composes specialized solvers  through problem-aware classification and dynamic orchestration, enabling robust handling of heterogeneous reasoning tasks.}
    \label{fig:framework}
\end{figure*}

Large language models (LLMs) have demonstrated sophisticated material inference properties but remain limited by hallucinations and logical inconsistencies \cite{zhang2023sirens,DeepSeekai2025DeepSeekr1incentivizingreasoningcapability,madusanka2025unravelling}. Formal reasoning systems, conversely, provide systematic guarantees with transparent and deterministic inference processes but require significant manual formalization \cite{nethercote2007minizinc, frederiksen2008pyke}. Recent advances in LLM-driven auto-formalization \cite{wu2022autoformalization,zhang2025formalizing} have enabled the development of  reasoning paradigms that bridge the material-formal gap \cite{pan2023logic,quan-etal-2025-peirce,xu-etal-2024-faithful,quan-etal-2024-verification,quan-etal-2024-enhancing,gorski2025integrating,quan-etal-2025-faithful}. However, existing approaches share a fundamental architectural constraint: they are designed for specific formal reasoning paradigms. For instance, LogicLLaMA \cite{yang2023harnessing} specializes in first-order logic translation, Sat-LM \cite{Ye-Et-Al:2023:SAT} targets Boolean satisfiability problems, LINC \cite{olausson2023linc} handles constraint satisfaction problems, and Logic-LM \cite{pan2023logic}, while supporting multiple logical solvers, requires specifying the reasoning strategy at design time. This static paradigm reflects an implicit assumption that the necessary strategies are known in advance, limiting the ability of neuro-symbolic/material-formal models to handle heterogeneous problems.

To address these challenges, we introduce a novel framework that enables a multi-paradigm adaptive reasoning. Our framework overcomes static solver integration through a context-aware reasoning strategy classification that automatically analyzes natural language inputs to identify required formal reasoning types in an unsupervised manner. Subsequently, the framework performs dynamic solver selection and composition, flexibly identifying appropriate formal logical solvers from an extensible library. 
This neuro-symbolic integration incorporates mechanisms to evaluate problem structure and adaptively select efficient formalization paradigms while maintaining expressiveness. 

We conduct an extensive empirical evaluation on five diverse benchmarks spanning logical reasoning, constraint satisfaction, and clinical trial-patient matching (ProntoQA, ProofWriter, FOLIO, LogDed, TREC$_{trials}$), alongside newly designed multi-question stress tests to evaluate solver selection and composition. Our findings support the following conclusions:

\begin{enumerate}
\item Dynamic reasoning strategy selection using LLMs can effectively enable adaptive behavior, with large frontier models achieving over 98\%  accuracy.
\item Adaptive solver composition supports effective neuro-symbolic inferences. On a mixed dataset requiring diverse strategies, our framework achieves 92.1\% accuracy, substantially outperforming zero-shot by 17.0\%, Chain-of-Thought by 21.4\%, and Symbolic CoT by 23.7\%.
\item Dynamic solver composition is critical for robustness on sequential reasoning tasks. When solving tasks containing sequences of heterogeneous problems, pure LLM methods achieve only 27.3\% overall accuracy while the proposed multi-paradigm framework achieves up to 54.4\%. 
This demonstrates that adaptive neuro-symbolic methods can offer a solution to substantially improve robustness on challenging reasoning problems which requires multi-paradigm formal inferences. 

\item Adaptive reasoning still poses challenges to smaller models, yet post-training enables significant improvements. In particular, small models show significant performance drops (8\% on Llama-3.1-8B, 12.6\% on Qwen-2.5-7B, 26.2\% on Qwen-2.5-Coder-7B), with 60-80\% of errors stemming from invalid formalization. However, fine-tuning using demonstrations from frontier models yields substantial improvements, reaching 41.0\% on Llama-3.1-8B, 50.1\% on Qwen-2.5-7B, and 59.6\% on Qwen-2.5-Coder-7B. 
\end{enumerate}

\section{Adaptive Multi-Paradigm Reasoning Framework}

Given a natural language reasoning problem $\boldsymbol{x}\in\mathcal{X}$ with ground truth answer $\boldsymbol{a}\in\mathcal{A}$, our goal is to construct an end-to-end framework $\mathcal{F}:\mathcal{X}\to\mathcal{A}$ that produces $\hat{\boldsymbol{a}} = \mathcal{F}(\boldsymbol{x})$ such that $\hat{\boldsymbol{a}} = \boldsymbol{a}$. Unlike traditional neuro-symbolic approaches that assume predetermined reasoning types, our framework handles raw natural language without prior knowledge of reasoning characteristics. As shown in Figure~\ref{fig:framework}, the framework consists of three main stages: (\textit{i}) problem decomposition, (\textit{ii}) routing and (\textit{iii}) reasoning, which formalizes the problem and calls the selected solver to produce predicted answers.

We thus formalize the end-to-end system  $\mathcal{F}:\mathcal{X}\to\mathcal{A}$ as a compositional map:
\begin{equation*}
\label{eq:F-factorization}
\begin{aligned}
\mathcal{F} \;=\;&
\mathsf{Reason}(\mathsf{Route}(\mathsf{Decompose}(\boldsymbol{x}))).
\end{aligned}
\end{equation*}

For all LLM-based modules, we utilize frontier models (GPT-4o, DeepSeek-V3) and three small-scale open-source models (Llama-3.1-8B, Qwen-2.5-7B, Qwen-2.5-Coder-7B), to assess the scalability of our approach, see Section \ref{sec:emp_setup} for details.

\paragraph{Problem Decomposition.}

Existing neuro-symbolic solver methods require rigid pre-specified problem types and structured inputs, which limits their practical applicability in scenarios where the reasoning strategy is not specified in advance.  Moreover, real-world applications may involve multi-reasoning scenarios in which problems need to be addressed by solving multiple sequential questions. To address these challenges, we leverage a LLM-based parser component responsible for structured problem decomposition.

The framework first converts $\boldsymbol x$ into a semi-structured representation capturing both decomposition and reasoning-type information. This step allows the system to handle inputs that may encompass multiple reasoning sub-tasks, each requiring a different formal paradigm:
\begin{equation*}
\label{eq:parser}
\begin{aligned}
\bigl(\mathcal{Q},\mathcal{T}\bigr) \;=\;& \mathsf{Decompose}(\boldsymbol{x}),\\
\mathcal{Q} \;=\;& \{Q_i\}_{i=1}^{n},\quad
\mathcal{T} \;=\; \{T_i\}_{i=1}^{n},\\
T_i \;\in \;&\{\mathrm{LP},\mathrm{FOL},\mathrm{CSP},\mathrm{SMT}, \ldots\}.
\end{aligned}
\end{equation*}
\noindent where each $Q_i = \{C_{i,j}\}_{j=1}^{m_i}$  represents the input \( \boldsymbol{x} \) structured into $m_i$ key components $C_{i,j}$  required for reasoning under the corresponding paradigm \( T_i \in \{\text{LP}, \text{FOL}, \text{CSP}, \text{SMT}, \ldots\} \), where LP stands for Logic Programming, FOL stands for First-Order Logic, CSP stands for Constraint Satisfaction Problems, and SMT stands for Satisfiability Modulo Theories.

\paragraph{Inference Routing.}

Given the decomposed representation $\mathcal{Q},\mathcal{T} = \mathsf{Decompose}(\boldsymbol{x})$, the \Router\ component determines which solver should handle each sub-problem based on its reasoning type. Formally, let
\[
\mathcal{S} = \{ S_T \mid T \in \{\mathrm{LP}, \mathrm{FOL}, \mathrm{CSP}, \mathrm{SMT}, \ldots \} \}
\]
denote the portfolio of available solvers. To leverage the reliability and interpretability of symbolic reasoning 
engines, we build each solver upon its corresponding symbolic reasoning engine (e.g., Pyke for LP and Prover9 for FOL, see Section \ref{sec:emp_setup} for details).

For each sub-question \( Q_i \in \mathcal{Q} \) with associated reasoning type \( T_i \in \mathcal{T} \), the router selects the corresponding solver \( S_{T_i} \) and invokes it to obtain a predicted answer:
\[
\hat{a_i} = S_{T_i}(Q_i).
\]

\paragraph{Reasoning: Autoformalization \& Solving.}
Each solver in the portfolio operates on inputs that strictly conform to its formal syntax, which is not compatible with natural language components $\mathcal{Q}$. To bridge this gap, we equip each solver with an LLM-based autoformalization preprocessing step. Specifically, given $Q_i = \{C_{i,j}\}_{j=1}^{m_i}$ with reasoning type $T_i$, the solver $S_{T_i}$ first employs an LLM to formalize each component $C_{i,j}$ into the syntax required by the reasoning engine for paradigm $T_i$. This representation is then fed into the symbolic reasoning engine to obtain inference results. Formally, the reasoning stage applies typed solvers via a formalize--solve pipeline:
\begin{equation*}
\label{eq:solver-pipeline}
\begin{aligned}
\hat{a}_i
&= S_{T_i}(Q_i)
\\
&= \texttt{Solve}_{T_i}(\texttt{Formalize}_{T_i}(Q_i, T_i))
\end{aligned}
\end{equation*}
\noindent where $\texttt{Formalize}(Q_i, T_i)$ translates each component $C_{i,j} \in Q_i$ into the formal language of solver $S_{T_i}$. The formalized problem is then solved with the symbolic engine. Next, the output is transformed into a final answer via an LLM. Finally, the outputs of the different sub-tasks are aggregated $\hat{\boldsymbol{a}} =  \{\hat{a}_i\}_{i=1}^{n}$.  
The prompts for autoformalization are presented in Appendix \ref{appen:sec_prompt_template}. Algorithm \ref{fig:algorithm} outlines the underlying end-to-end process, and Appendix \ref{appen:plan_execution} the implementation details for the Agentic workflow execution.

\begin{figure}[!t]
	\renewcommand{\algorithmicrequire}{\textbf{Input:}}
	\renewcommand{\algorithmicensure}{\textbf{Output:}}
    \small
	\begin{algorithm}[H]
		\caption{Overall Workflow}
		\label{gpv}
		\begin{algorithmic}
			\REQUIRE Natural Language Input $\boldsymbol x$
			\ENSURE Final Answer $ \hat{\boldsymbol{a}} $
            \STATE \texttt{/Stage 1|Problem Decomposition/}
			\STATE $\mathcal{Q}, \mathcal{T} = \texttt{Parser} (\boldsymbol{x})$
			\STATE \texttt{/Stage 2|Solver Registration/}
			\STATE $\mathcal{S} = \texttt{Register}()$
            \STATE \texttt{/Stage 3|Dynamic Reasoning Composition/}\\
            \STATE $\hat{\boldsymbol{a}} = [ \ \ ]$
            \FOR{$Q \in \mathcal{Q}, T \in\mathcal{T}$}
            \STATE $S_T = \texttt{Router}(T)$  \quad  \texttt{// Solver Selection}\\
            \STATE $\boldsymbol{l} = \texttt{Formalize}_{T}(Q)$\texttt{// Formalization}
            \STATE $\hat{a} = \texttt{Solve}_{T}(\boldsymbol{l})$ \quad\texttt{// Calling Solver}
            \STATE $\hat{\boldsymbol{a}}\texttt{.append}(\hat{a})$
            \ENDFOR
		\end{algorithmic}
    \label{fig:algorithm}
	\end{algorithm}
\end{figure}

\section{Empirical Analysis}

\begin{table*}[!ht]
  \small
    \centering
    \setlength {\tabcolsep }{1mm}
    \resizebox{\textwidth}{!}{
      \begin{tabular}{c|cc|cc|cc|cc|cc|cc|c}
      \toprule
      \multirow{2}[0]{*}{\textbf{Methods}} & \multicolumn{2}{c|}{\textbf{PrOntoQA}} & \multicolumn{2}{c|}{\textbf{ProofWriter}} & \multicolumn{2}{c|}{\textbf{FOLIO}} & \multicolumn{2}{c|}{\textbf{LogDed7}} & \multicolumn{2}{c|}{\textbf{TREC}$_{trials}$} & \multicolumn{2}{c|}{\textbf{Mixed}} & \multirow{2}[0]{*}{\textbf{Routing}} \\
            & w/o routing & w/ routing & w/o routing & w/ routing & w/o routing & w/ routing & w/o routing & w/ routing & w/o routing & w/ routing & w/o routing & w/ routing &  \\
      \midrule
    \multicolumn{14}{c}{\textbf{GPT-4o}} \\
    \midrule
    Zero-shot & 82.2  & $\underline{91.6}$ & 45.5  & $\underline{60.3}$ & 63.7  & $\underline{\textbf{76.0}}$ & 67.3  & $\underline{74.4}$ & 71.3  & $\underline{78.0}$ & 65.1  & $\underline{75.1}$ & \multirow{4}{*}{{98.0}}  \\
    CoT   & 84.4  & $\underline{90.2}$ & 43.5  & $\underline{52.5}$ & 64.7  & $\underline{71.1}$ & 67.9  & $\underline{71.0}$ & 70.0  & $\underline{73.7}$ & 65.1  & $\underline{70.7}$ &    \\
    CoT$_{sym}$ & 75.4  & $\underline{97.8}$ & 44.0  & $\underline{53.7}$ & 66.7  & 53.4  & 63.6  & 61.4  & 67.7  & $\underline{75.7}$ & 61.8  & $\underline{68.4}$ &   \\
    Ours  & -     & \textbf{99.4} & -     & \textbf{94.7} & -     & 69.1  & -     & \textbf{95.6} & -     & \textbf{82.3} & -     & \textbf{92.1} &  \\ \midrule
    \multicolumn{14}{c}{\textbf{DeepSeek-V3.1}} \\
\midrule
Zero-shot & 7.8 & 2.2 & 7.0 & 3.8 & 17.2 & $\underline{18.1}$ & 36.0 & 14.7 & 46.7 & 42.3 & 22.0 & 13.1 & {\multirow{4}{*}{99.3}} \\
CoT & 24.8 & 2.0 & 18.3 & 2.2 & 44.6 & 28.4 & 57.0 & 37.7 & 27.0 & 17.7 & 34.9 & 17.3 & \textbf{ } \\
CoT$_{sym}$ & 57.8 & $\underline{91.2}$ & 37.8 & $\underline{57.0}$ & 66.2 & $\underline{\textbf{73.5}}$ & \textbf{64.4} & 51.7 & 66.7 & $\underline{\textbf{79.7}}$ & 56.5 & $\underline{67.2}$ & \textbf{ } \\
Ours & - & \textbf{97.6} & - & \textbf{81.3} & - & 52.0 & - & 57.7 & - & 68.0 & - & \textbf{73.4} & \textbf{ } \\
    \bottomrule
  \end{tabular}%
  }
  \caption{Overall \texttt{pass@1} performance on representative proprietary and open-source models. ``w/o routing'' and ``w/ routing'' denote the performance without and with the auxiliary reasoning selection strategy. {Bold} indicates the best setting for each dataset. {Underline} indicates improvement after adding routing information. The last column reports the Router's prediction accuracy on the Mixed dataset. Since all settings use the same prompt and planning accuracy shows minimal variation across different settings, we report the mean accuracy for each backbone model.}
  \label{tab:empirical_evaluation}
\end{table*}%

\subsection{Experimental Settings} 
\label{sec:emp_setup}
\paragraph{Datasets.} In order to evaluate the proposed dynamic reasoning framework, we select a heterogeneous set of datasets spanning logical reasoning, constraint satisfaction and clinical trial-patient matching including: {PrOntoQA} \cite{saparov2023prontoqa}, {ProofWriter} \cite{tafjord-etal-2021-proofwriter}, {FOLIO} \cite{han-etal-2024-folio}, {LogDed7} \cite{srivastava2023beyond}, {TREC}$_{trials}$ \cite{soboroff2021overviewTREC}, where {PrOntoQA} and {ProofWriter} represent LP problems, {FOLIO} represents FOL problems, {LogDed7} represents CSP problems and  {TREC}$_{trials}$ represents SMT problems. Additional details on the datasets and their formal selection criteria can be found in Appendix \ref{append:datasets}. We convert all instances from the five datasets into a unified natural language format using task-specific templates to simulate the pure natural language input scenarios encountered in real-world applications (see Appendix \ref{appen:prompt_for_merge}), then shuffle and merge them into a single \textbf{Mixed} dataset. Under this setup, the framework must first identify the reasoning strategy, then compose the relevant solvers and perform reasoning accordingly.

\paragraph{Baselines.} We use LLMs under different prompting settings as baselines. Specifically, we evaluate three LLM settings, including: Zero-shot, Chain-of-Thought (CoT) \cite{kojima2022large}, Symbolic CoT (CoT$_{sym}$) \cite{xu-etal-2024-faithful}. Additionally, to validate the effectiveness of the routing
mechanism within our framework, we integrate the Router’s outputs as auxiliary prompts into the aforementioned Zero-shot, CoT, and CoT$_{sym}$
settings. We also analyse the benefits of adaptive reasoning over static neuro-symbolic approaches (the paradigm of previous methods such as Logic-LM \cite{pan2023logic} or VERUS-LM \cite{callewaert2025verus}) through dedicated ablation studies in Section 3.4.

\begin{table*}[!ht]
  \small
    \centering
    \setlength {\tabcolsep }{1mm}
    \resizebox{\textwidth}{!}{
      \begin{tabular}{c|cc|cc|cc|cc|cc|cc|c}
      \toprule
      \multirow{2}[0]{*}{\textbf{Methods}} & \multicolumn{2}{c|}{\textbf{PrOntoQA}} & \multicolumn{2}{c|}{\textbf{ProofWriter}} & \multicolumn{2}{c|}{\textbf{FOLIO}} & \multicolumn{2}{c|}{\textbf{LogDed7}} & \multicolumn{2}{c|}{\textbf{TREC}$_{trials}$} & \multicolumn{2}{c|}{\textbf{Mixed}} & \multirow{2}[0]{*}{\textbf{Routing}} \\
            & w/o routing & w/ routing & w/o routing & w/ routing & w/o routing & w/ routing & w/o routing & w/ routing & w/o routing & w/ routing & w/o routing & w/ routing &  \\
      \midrule
    \multicolumn{14}{c}{\textbf{Llama-3.1-8b}} \\
    \midrule
Zero-shot & 65.2 & 58.6 & 41.8 & 36.3 & \textbf{46.7} & 42.6 & \textbf{37.0} & 25.3 & 63.2 & 39.7 & \textbf{48.7} & 38.8 & {\multirow{5}{*}{76.8}} \\
CoT & 57.9 & 54.1 & 45.3 & 35.4 & 38.2 & $\underline{42.3}$ & 19.5 & $\underline{24.0}$ & \textbf{63.6} & 48.3 & 42.0 & 38.3 & \textbf{ } \\
CoT$_{sym}$ & 10.1 & 0.0 & 10.8 & 0.2 & 12.1 & 0.2 & 3.9 & 0.6 & 2.6 & $\underline{3.1}$ & 7.6 & 0.7 & \textbf{ } \\
Ours & - & 3.9 & - & 5.3 & - & 0.8 & - & 7.0 & - & 27.6 & - & 8.0 & \textbf{ } \\
Ours$_{sft}$ & - & \textbf{75.7} & - & \textbf{51.5} & - & 31.5 & - & 10.1 & - & 40.9 & - & 41.0 & \textbf{ } \\
    \midrule
    \multicolumn{14}{c}{\textbf{Qwen-2.5-7b}} \\
    \midrule
Zero-shot & 57.2 & 40.9 & 37.8 & 29.8 & 32.4 & $\underline{40.0}$ & 42.9 & 38.5 & 59.3 & 53.2 & 45.9 & 38.8 & {\multirow{5}{*}{97.7}} \\
CoT & \textbf{95.3} & 72.8 & \textbf{59.8} & 59.1 & 48.0 & 47.7 & 46.8 & 20.6 & 47.0 & 25.3 & \textbf{60.8} & 45.0 & \textbf{ } \\
CoT$_{sym}$ & 55.1 & 46.1 & 41.7 & 29.7 & \textbf{59.2} & 52.3 & 8.0 & $\underline{\textbf{49.5}}$ & 60.2 & $\underline{\textbf{61.9}}$ & 38.3 & $\underline{45.5}$ & \textbf{ } \\
Ours & - & 0.0 & - & 0.0 & - & 0.0 & - & 31.5 & - & 23.1 & - & 12.6 & \textbf{ } \\
Ours$_{sft}$ & - & 71.8 & - & 51.9 & - & 6.4 & - & 45.0 & - & 51.9 & - & 50.1 & \textbf{ } \\
    \midrule
    \multicolumn{14}{c}{\textbf{Qwen-2.5-Coder-7b}} \\
    \midrule
Zero-shot & 71.5 & 67.8 & 53.6 & $\underline{\textbf{56.5}}$ & 45.3 & 25.7 & 45.8 & 44.1 & 66.2 & $\underline{\textbf{66.7}}$ & 56.0 & 53.8 & {\multirow{5}{*}{92.9}} \\
CoT & 70.5 & 64.8 & 51.3 & $\underline{53.3}$ & 45.4 & 24.8 & 40.5 & 40.4 & 63.0 & $\underline{65.2}$ & 53.2 & 50.9 & \textbf{ } \\
CoT$_{sym}$ & 78.2 & 29.8 & 46.8 & 20.6 & \textbf{52.1} & 13.2 & 14.0 & 11.3 & 38.0 & 29.3 & 43.0 & 20.3 & \textbf{ } \\
Ours & - & 6.1 & - & 12.8 & - & 2.8 & - & 60.2 & - & 23.2 & - & 26.2 & \textbf{ } \\
Ours$_{sft}$ & - & \textbf{89.4} & - & 44.9 & - & 3.1 & - & \textbf{74.2} & - & 43.3 & - & \textbf{59.6} & \textbf{ } \\
    \bottomrule
  \end{tabular}%
  }
  \caption{Average performance of small open-source models over three random seeds.
  Since all settings use the same routing result from the pretrained model, we only report the accuracy once under the same backbone model.}
  \label{tab:empirical_evaluation_small}
\end{table*}%

\paragraph{Foundation Models.} We use four reference foundation models across different families to evaluate the proposed approach: one large proprietary model ({GPT-4o} \cite{openai2024gpt04o}), one large open-source model ({DeepSeek-V3.1} \cite{DeepSeekai2024DeepSeekv3technicalreport}) and three smaller models ({Llama-3.1-8B-Instruct} \cite{grattafiori2024llama}, {Qwen-2.5-7B-Instruct} \cite{qwen2024qwen205}, and {Qwen-2.5-Coder-7B-Instruct} \cite{qwen2024qwen205}), under Zero-shot, CoT, CoT$_{sym}$ settings.  
For smaller models, we use the pretrained versions for problem decomposition and routing in our framework as they demonstrate stronger performance. For the Solver component, we employ pretrained versions for autoformalization. Additionally, we experiment with fine-tuning the small models on formalizations generated by GPT-4o to investigate the potential for performance improvement via distillation methods.

To manage inference costs, we evaluate {GPT-4o} and {DeepSeek-V3.1} using pass@1 metric, measuring accuracy based on a single inference run.
For small models, we conduct experiments three times with different seeds and report the average performance across three runs. Additional implementation details can be found in Appendix \ref{append:eval_details}.

\paragraph{Logical Solvers.} We instantiate our framework with a set of symbolic {solvers} to support logical reasoning tasks. Four back-end engines are integrated: {Pyke} \cite{frederiksen2008pyke} for LP problems, {Prover9} \cite{prover9-mace4} for FOL problems, Z3 \cite{10.1007/978-3-540-78800-3_24} for  SMT problems, and MiniZinc \cite{nethercote2007minizinc} for CSP problems, respectively. Pyke performs rule-based inference over formalized premises and hypotheses, while Prover9 conducts first-order deduction. Z3 checks satisfiability of symbolic constraints within logical theories, and MiniZinc solves combinatorial search problems over variable domains to produce concrete assignments. See Appendix \ref{appen:solvers} for further details.

\subsection{Results on Frontier Models}
\label{sec:res_on_frontier}
Table \ref{tab:empirical_evaluation} compares the performance before and after incorporating the multi-reasoning planning, under different foundation models. In addition, the last column reports the routing accuracy, i.e., the proportion of test cases where the Router correctly selects the expected solver.
Based on the results in Table \ref{tab:empirical_evaluation}, we draw the following conclusions:

\paragraph{The proposed adaptive neuro-symbolic framework achieves the best overall performance compared to all baselines.}
Overall, our framework combined with larger models achieves the best performance, reaching 92.1\% accuracy on the Mixed dataset. Our framework also yields significant improvements across different reasoning types, which demonstrate the superior performance of combining LLM with symbolic solvers.

\paragraph{Dynamic routing strategy is effective using large frontier models.} As shown in the last column, both GPT-4o and DeepSeek-V3.1 achieve routing accuracy exceeding 98\%, with DeepSeek-V3.1 even slightly outperforming GPT-4o. This demonstrates that current frontier models are capable of dynamically 
parsing and identifying reasoning types, which validates the feasibility of the proposed dynamic routing strategy.

\paragraph{Dynamic routing strategy is effective on improving reasoning performance on frontier models.} 
Comparing performance with and without reasoning strategy selection under the same setting, we observe that the dynamic strategy significantly 
improves model performance in most cases, particularly for GPT-4o. This improvement is especially pronounced for the CoT-Sym method, which, leveraging formalization for reasoning. Using adaptive reasoning, this method shows substantial gains on both GPT-4o and DeepSeek-V3.1 (e.g., 33.4\% improvement on ProntoQA for DeepSeek-V3.1 and 22.4\% for GPT-4o). This demonstrates that the dynamic routing strategy not only plays a crucial role in enabling adaptive reasoning within a neuro-symbolic framework, but also serves as effective guidance that can enhance the reasoning capabilities of language models themselves.

\paragraph{Reasoning strategy and model size affect formalization and reasoning performance.}
Although our framework demonstrates strong performance with both GPT-4o and DeepSeek-V3.1, DeepSeek-V3.1 lags noticeably behind GPT-4o despite its better routing accuracy. This performance gap stems primarily from differences in formalization quality (see Section \ref{sec:error_ana} for detailed analysis), which is influenced by both reasoning type complexity and model capability. The impact of reasoning type is evident in the FOLIO results in Table \ref{tab:empirical_evaluation}, where both models perform significantly worse on FOLIO (FOL problems) compared to ProntoQA and ProofWriter (LP problems), despite all involving deductive reasoning. This can be explained by the difference in formalizing first-order vs propositional logic. The effect of model size on formalization is observable not only in the comparison between GPT-4o and DeepSeek-V3.1, but will be further demonstrated in the next section through experiments with smaller open-source models.

\subsection{Results on Smaller Models}
\label{sec:res_on_small}
We experimented with small-scale open-source models to examine their performance and limitations in adaptive reasoning, with results shown in Table \ref{tab:empirical_evaluation_small} (see Appendix \ref{appen:complete_res} for complete results with standard deviations). We observe that the dynamic strategy selection maintains strong performance on small-scale models, achieving over 90\% accuracy on Qwen-series models and 76.8\% on Llama-3.1-8B. Overall, the neuro-symbolic framework achieves the best performance using Qwen-2.5-Coder-7B. However, a notable performance gap exists compared to closed-source models, supporting our earlier conclusion on model size's effect on autoformalization. Notably, while Qwen2.5-7B's Routing Accuracy (97.7\%) nearly matches GPT-4o (98.0\%), its reasoning performance lags significantly behind. Upon further analysis, we find that:

\paragraph{Pre-trained smaller models lack structural abstraction over formal languages necessary for external solver integration.}
Recall that the LP solver, FOL solver, CSP solver, and SMT solver are implemented in Prover9, Pyke, MiniZinc, and Z3, all 'under-resourced' formal languages. Without knowledge of the target syntax, small models often produce invalid formal representations. 

\paragraph{However, performance of smaller models can be significantly improved via supervised fine-tuning.} Despite fine-tuning with only 5K samples, we observe significant performance improvements across all models (33\% for Llama-3.1-8B, 37.5\% for Qwen-2.5-7B, and 33.4\% for Qwen2.5-Coder-7B). This further demonstrates the critical impact of autoformalization quality on our framework's capabilities and highlights persisting deficiencies of small-scale models for effective neuro-symbolic reasoning.

\begin{table}[!t]
  \small
    \centering
    \setlength {\tabcolsep }{1mm}
    \resizebox{0.8\linewidth}{!}{
      \begin{tabular}{c|cc}
      \toprule
      \multirow{2}[0]{*}{\textbf{Models}} &  \multicolumn{2}{c}{\textbf{Mixed}}  \\
      & random routing & w/ routing   \\
\midrule
GPT-4o & 29.0 & ${\textbf{92.1}}$ \\
DeepSeek-V3.1  & {32.8} & ${73.4}$ \\
Llama-3.1-8b & 26.0 & ${41.0}$ \\
Qwen-2.5-7b & 29.6 & ${50.1}$ \\
Qwen-2.5-Coder-7b & 20.5 & ${59.6}$ \\
\bottomrule
\end{tabular}
}
  \caption{Ablation performance with random routing.}
  \label{tab:ablation}
\end{table}
\subsection{Ablation Study on Adaptive Reasoning}
A core feature of the adaptive neuro-symbolic framework lies in the ability to dynamically select and integrate an appropriate solver for reasoning through the Router. To demonstrate the effectiveness of this component, we ablate it by randomly assigning solvers to problems. Note that in a random solver setting, the decomposed problem components may be incompatible with the assigned solver (e.g., a CSP solver cannot extract the required constraints from an LP problem). Therefore, we allow the solver to directly formalize the original input text. We evaluate GPT-4o, DeepSeek-V3.1, and three small models. For each model, we conduct experiments with 3 random seeds, with average results shown in Table \ref{tab:ablation}. The ablations demonstrate that the dynamic routing mechanism is fundamental for effective neuro-symbolic integration, and ablating it leads to significant performance degradation across all models, which demonstrates the importance of incorporating adaptive reasoning strategies for addressing heterogeneous problems.

\begin{figure}[!t]
    \centering
    \includegraphics[width=\linewidth]{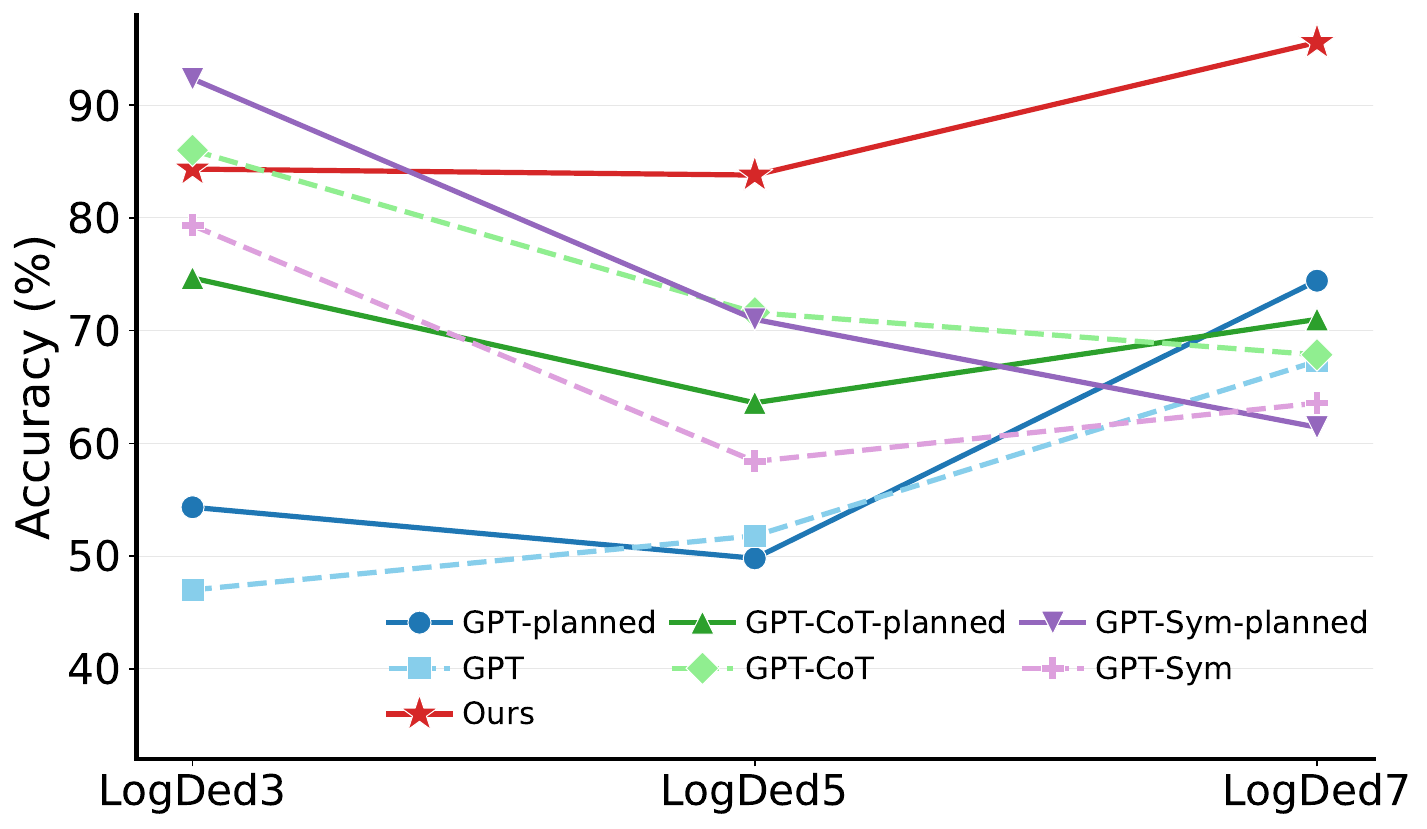}
    \caption{Model performance on Logic Deduction tasks of varying difficulty.} 
    \label{fig:csp}
\end{figure}
\begin{figure}[!t]
    \centering
    \includegraphics[width=\linewidth]{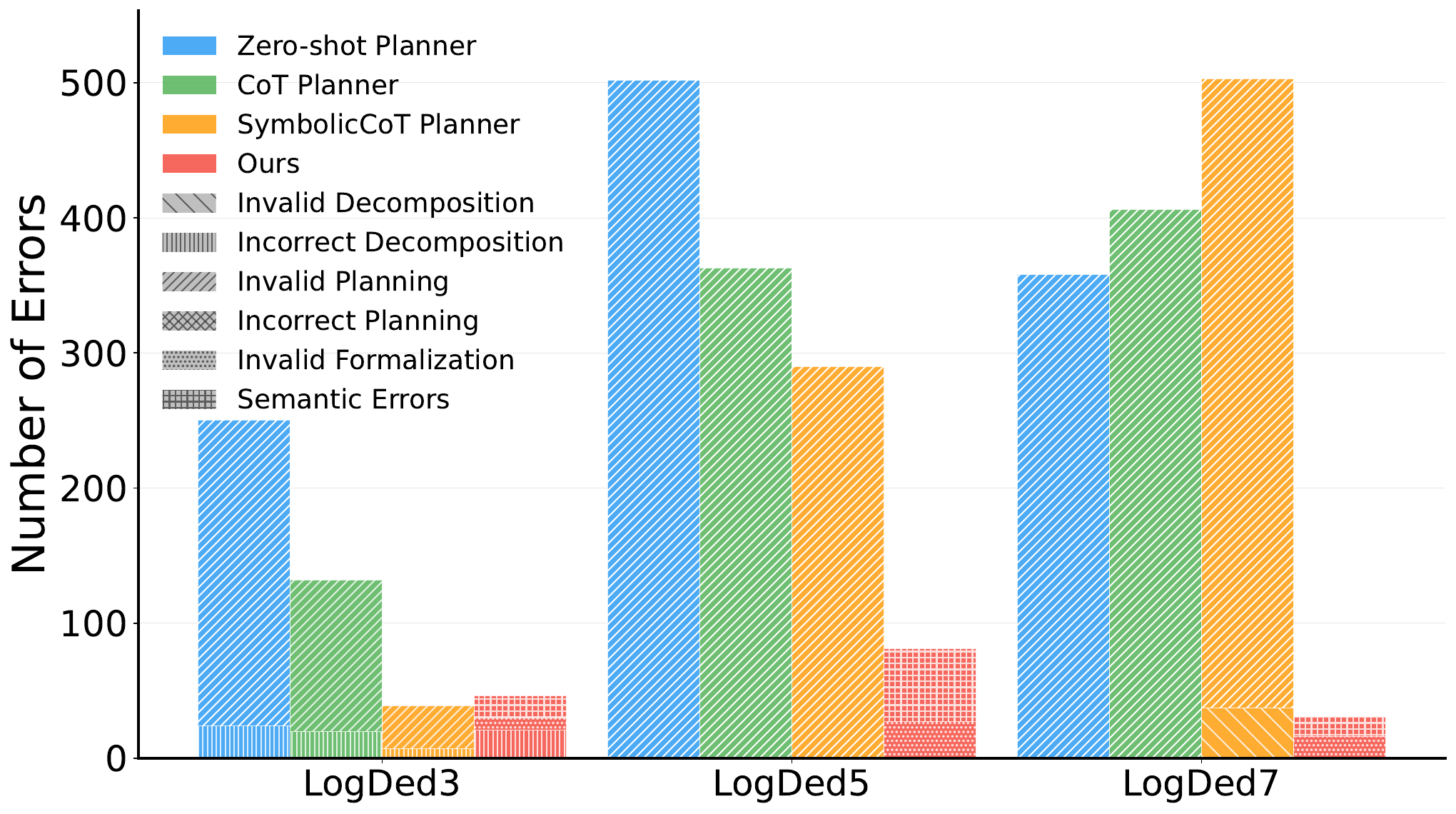}
    \caption{ Distribution of error types of our framework for Logic Deduction tasks.}
    \label{fig:error_ana_csp}
\end{figure}

\subsection{Performance Across Varying Difficulty Levels}

To investigate how the framework performs across problems of varying difficulty levels, we selected the Logical Deduction subset from BigBench \cite{DBLP:journals/tmlr/SrivastavaRRSAF23}, which requires deducing the order of a sequence of objects from a minimal set of conditions and contains an increasing number of variables (i.e. 3, 5, and 7, denoted as LogDed3, LogDed5, and LogDed7, respectively). These subsets form an evaluation suite with increasing levels of reasoning complexity. Given GPT-4o’s strong capabilities in formalization and logical understanding, we choose it as the backbone model to examine how different methods perform under varying logical deduction difficulty.

Figure~\ref{fig:csp} shows that our framework achieves the best overall performance across all difficulty levels. Furthermore, our method exhibits an unexpected upward trend in accuracy as the task becomes more difficult. 
An investigation on the error distribution (detailed in Section \ref{sec:error_ana}) in Figure~\ref{fig:error_ana_csp} shows that 
all settings exhibit a substantial number of \textit{Incorrect Decomposition} cases on LogDed3, whereas such errors are rare on LogDed5 and LogDed7.  We hypothesize that the lower number of variables and the resulting shorter problem formulation provide less contextual information for identifying the task as a CSP problem, making it harder for the Router to make accurate decisions. 

\subsection{Performance in Multi-Question Scenarios}
Real-world scenarios often present composite queries with sub-problems requiring different reasoning paradigms.
In this section, we evaluate our proposed framework in a more complex, compositional setting, where models are required to sequentially solve multiple reasoning problems. Inspired by the experimental setup in Reasoning Evaluation through Simultaneous Testing (REST) \cite{pan2025reststresstestinglarge}, we extend our evaluation to require the solution of multiple reasoning problems in a single pass. 

Specifically, we randomly shuffle the Mixed dataset and group the data into batches of three examples, concatenating each group into a single input prompt for joint inference. We define \textit{Overall Accuracy} to measure global-level reasoning consistency, which requires all three answers to be correct. Details on the construction of this evaluation setting are described on Appendix \ref{appen:sec_prompt_template}. GPT-4o is used as the reference model for this evaluation given its performance on the mixed dataset.

Table~\ref{tab:concat} shows the performance of all settings for the multi-question task, where we find that:

\paragraph{Explicit problem decomposition is essential for composite reasoning tasks.} Our framework outperforms the second-best method by 16.2\% in \textit{Overall Accuracy}, demonstrating the critical value of systematic parsing and planning for composite reasoning tasks.

\paragraph{Routing integration provides universal benefits for cross-task consistency.} 
Consistent \textit{Overall Accuracy} improvements across all Router-integrated methods (2.5\% for GPT-Zero-shot, 0.7\% for GPT-CoT, and 17.2\% for GPT-CoT$_{sym}$) reveals the impact of joint decomposition and coordination across different reasoning types, which prevent reasoning drift across tasks. This suggests that maintaining logical coherence across composite tasks requires explicit orchestration lacking in monolithic approaches.

\begin{table}[!t]
\small
  \centering
  \resizebox{0.8\linewidth}{!}{
    \begin{tabular}{c|cc}
    \toprule
    \multicolumn{1}{c|}{\multirow{2}[2]{*}{\textbf{Methods}}} & \multicolumn{2}{c}{\textbf{Overall Acc.}} \\
    \multicolumn{1}{c|}{} & w/o routing & \multicolumn{1}{c}{w/ routing}  \\
    \midrule
    \textbf{GPT-Zero-Shot} & \multicolumn{1}{c}{24.5 } & $\underline{27.0}$   \\
    \textbf{GPT-CoT} & \multicolumn{1}{c}{27.3 } & $\underline{28.0}$   \\
    \textbf{GPT-CoT}$_{Sym}$ & \multicolumn{1}{c}{21.0 } & $\underline{38.2}$  \\
    \textbf{Ours} & -     & \textbf{54.4} \\
    \bottomrule
    \end{tabular}%
    }
  \caption{Performance on the sequential multi-question reasoning task. {Bold} indicates the best result for each dataset. {Underline} indicates improvement after adding routing information.}
  \label{tab:concat}
\end{table}%

\subsection{Error Analysis}
\label{sec:error_ana}
For each input problem, our framework proceeds through five stages: (1) problem decomposition, (2) routing, (3) autoformalization, (4) symbolic reasoning, and (5) output translation. Since reasoning relies on deterministic symbolic reasoners with formal rules, we observe errors primarily under the generative-dominated stages (1), (2), (3) and (5).

Specifically, we categorize the failure cases into the following six types: 1. \textbf{Invalid Decomposition}: Invalid problem decomposition or reasoning type classification. 2. \textbf{Incorrect Decomposition}: Misclassification of the problem type. 3. \textbf{Invalid routing}: Invalid or empty output generated by the Router, causing failure in composing solvers. 4. \textbf{Incorrect routing}: Using inappropriate solver for solution. 5. \textbf{Invalid Formalization}: Formal code that does not conform to the target solver’s syntax, causing the reasoner to fail. 6. \textbf{Semantic Errors}: All remaining incorrect predictions that do not fall into the above five categories. Semantic errors typically arise from two sources: (i) the LLM generates syntactically valid but semantically incorrect formal code (e.g., wrong variable assignments or misaligned rules), leading the solver to an incorrect conclusion; or (ii) the LLM mistranslates the solver’s output during the final decoding step, mapping it to the wrong label.

\begin{figure}[!t]
    \centering
    \includegraphics[width=\linewidth]{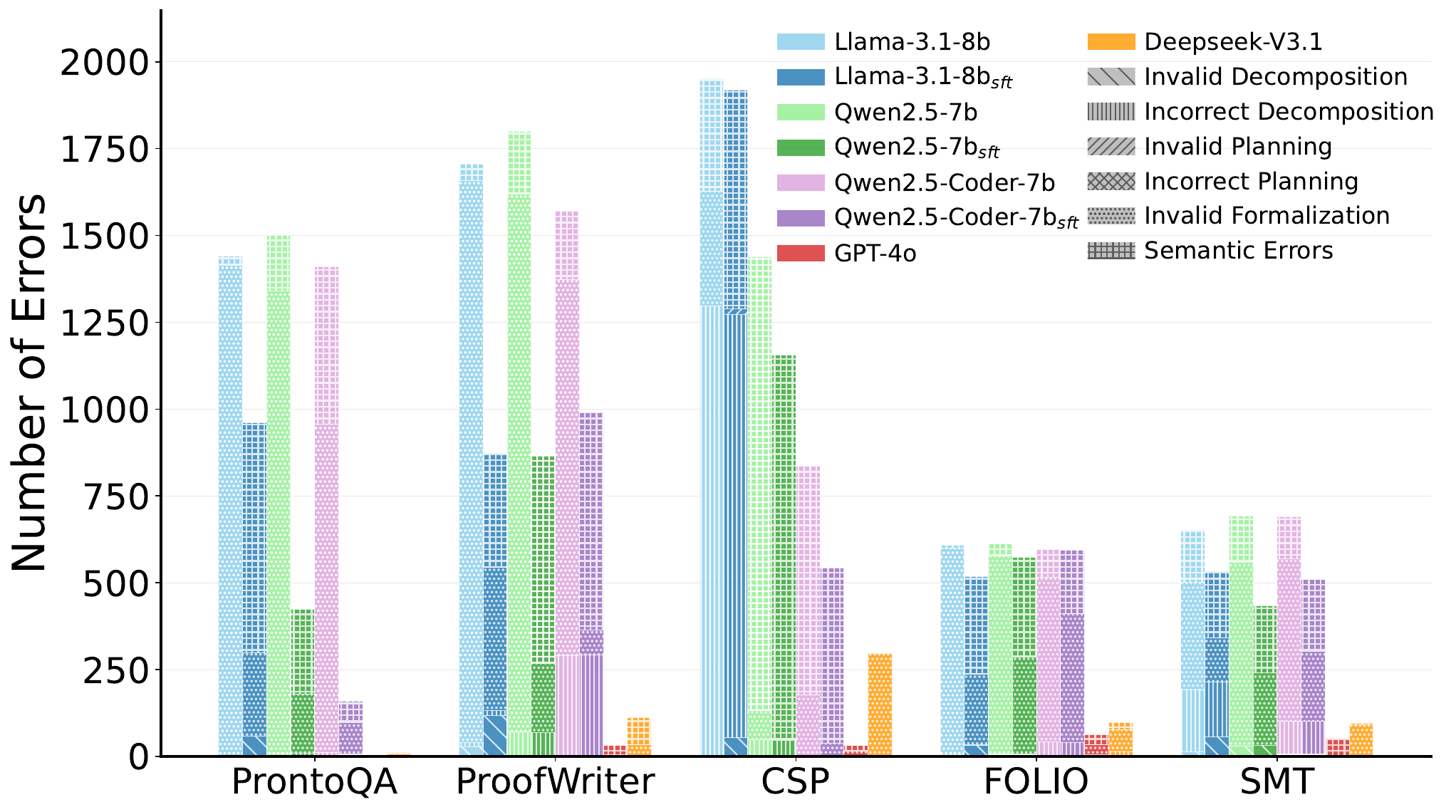}
    \caption{Distribution of error types of our framework. Each bar represents the number of incorrect predictions made by a model on a given dataset. Within each bar, different fill patterns indicate different error types.}
    \label{fig:error_ana}
\end{figure}

\paragraph{Error taxonomy reveals model-specific failure regimes: small models collapse early but can be improved via fine-tuning, GPT-4o fails post-reasoning.}
We report error type distributions of different models in our framework in Figure~\ref{fig:error_ana}. As noted in Section \ref{sec:res_on_frontier}, open-source models demonstrate limited formalization capabilities, consequently degrading system performance. Figure~\ref{fig:error_ana} reveals that \textit{Invalid Formalization} constitutes the predominant error type for open-source models (including DeepSeek-V3), suggesting systematic failures in producing syntactically valid symbolic representations, particularly for domain-specific formal languages. However, such errors are significantly reduced via fine-tuning. In contrast, GPT-4o’s errors are primarily categorized as \textit{Semantic Errors}, with only a few cases of \textit{Invalid Formalization} observed on the FOLIO and CSP datasets.

\section{Related Work}
Recent work increasingly combines large language models with symbolic solvers to improve logical reasoning. Early approaches focused on translating natural language into specific formal representations: LogicLLaMA \cite{yang2023harnessing} specializes in NL-to-FOL translation while Sat-LM \cite{Ye-Et-Al:2023:SAT} targets NL-to-SAT translation. To address formalization uncertainty, LINC \cite{olausson2023linc} generates multiple SAT formalizations and applies majority voting on outputs. Building on these foundations, Logic-LM \cite{pan2023logic} supports multiple logical languages and solvers, while VERUS-LM \cite{callewaert2025verus} adds self-refinement via feedback from reasoning engines. Large-scale evaluations by \citet{jiang2025large} demonstrate that smaller models can handle various formalizations when trained on curated data.

Further advancing compositionality, CLOVER \cite{ryudivide} decomposes NL into atomic substatements with logical dependencies, and $\forall$uto$\exists$val \cite{karia2024forallutoexistsvalautonomousassessmentllms} ensures FOL formalization fidelity through bidirectional NL-to-formal syntax mapping. More recent work embeds symbolic structure directly into LLM reasoning. Methods such as SymbCoT \cite{xu2024faithful} and QuaSAR \cite{ranaldi-etal-2025-improving} integrate symbolic expressions into CoT prompting for improved faithfulness, while PEIRCE \cite{quan-etal-2025-peirce} introduces a neuro-symbolic refinement loop where LLM-generated hypotheses are iteratively critiqued by symbolic provers and soft evaluators.

\section{Conclusion}
In this paper, we present an adaptive and extensible framework for end-to-end LLM-supported symbolic reasoning. By decomposing the reasoning process into three explicit stages—problem decomposition, routing, and solver-based reasoning, our framework provides a unified interface for handling a diverse range of reasoning tasks. Its agent-based design allows for a dynamic and extensible interface of reasoning functions. Auto-formalization cycles, closely connected to a set of symbolic solvers ensure that reasoning is formally grounded, enabling both interpretability and epistemic trust. 

\section*{Limitations}

While our framework demonstrates promising results for adaptive LLM-symbolic reasoning, several limitations warrant discussion:

\paragraph{Model Scale Dependencies.} Our experiments reveal a strong dependency on model scale for effective symbolic integration. Pre-trained smaller models (7B-8B parameters) show significant performance degradation when integrated with external solvers, achieving only 8.0\% accuracy on mixed datasets compared to 92.1\% with GPT-4o. This limitation suggests that effective LLM-symbolic reasoning may be constrained to frontier models or require substantial fine-tuning investments for smaller models as we have explored in this paper.

\paragraph{Formalization Bottleneck.} The framework's performance is fundamentally limited by the quality of auto-formalization. Invalid formalization constitutes the primary error source for smaller models, indicating that the translation from natural language to formal representations remains a critical bottleneck. While our results show that fine-tuning can significantly improve formalization capabilities (improving performance from 8.0\% to 41.0\% on Llama-3.1-8b), this dependency on formalization quality may still limit the framework's applicability to domains with complex or non-standard logical structures.  Few-shot learning approaches or domain-specific fine-tuning strategies are ways to address this limitation.

\paragraph{Limited Solver Coverage.} While our framework supports four solver types (LP, FOL, CSP, SMT), it does not cover the full spectrum of formal reasoning paradigms. Mathematical reasoning, probabilistic inference, temporal logic, and other specialized reasoning domains are not addressed. However, the proposed model is extensible to additional solvers as new reasoning agents can be added to the portfolio without modifying the core routing framework. This extensibility represents a key design principle that should enable future expansion to other reasoning domains as new solvers become available.

\section*{Acknowledgements}
This work was partially funded by the Swiss National Science Foundation (SNSF) projects RATIONAL  and M-RATIONAL.


\bibliography{custom}

\appendix

\section{Solvers Used in Our Framework}
\label{appen:solvers}

Our framework includes a set of specialized, instantiated agents for formal inference. All formal agents are derived from a common base class, \texttt{LLMToolAgent}, which is instantiated with a specific backend solver and associated methods for \textit{formalizing}, \textit{reasoning}, and \textit{converting}. The \textit{formalizing} method translates the natural language problem into a formal representation compatible with the syntax of the embedded backend reasoning engine. The solver then performs \textit{reasoning} over this formalized input using the backend engine. Finally, the \textit{conversion} method maps the reasoning output back to the corresponding option label among the candidate answers provided in the original problem. Our solvers fall into two main categories: solvers for logical deduction and solvers for constraint satisfiability.

\subsection{Solvers for Logical Deduction} 
Our framework includes a set of pre-defined instantiated {solvers} to support logical reasoning tasks. Using the controlled autoformalization method described in PEIRCE \cite{quan-etal-2025-peirce}, each solver employs LLM-based auto-formalization to convert natural language problems into formal representations, with iterative refinement capabilities to automatically correct syntax errors when the generated code returns an error message. 
Two back-end solvers are implemented in our framework: \textit{Pyke} \cite{frederiksen2008pyke} and \textit{Prover9} \cite{prover9-mace4} as backend-solvers, allowing \textbf{Logic Programming (LP)} and \textbf{First-order Logics (FOL)} respectively, a set which can be expanded by instantiating the \texttt{LLMToolAgent} interface.

\subsection{Solvers for Constraint Satisfiability} 
Two other solver types are implemented: \textbf{Satisfiability Modulo Theories (SMT)} and \textbf{Constraint Satisfaction Problems (CSP)} solvers. The SMT Solver uses the \textit{Z3} theorem, auto-formalizing natural language constraints into the SMT-LIB format, then querying \textit{Z3} to determine satisfiability. This approach is particularly effective for matching problems involving integer constraints, such as determining patient eligibility for clinical trials based on age, disease stage, and other medical criteria. The CSP Solver employs the \textit{MiniZinc} constraint modeling language \cite{nethercote2007minizinc} and the \textit{Gecode} solver to handle Constraint Satisfaction Problems, also auto-formalizing natural language problem descriptions into \textit{MiniZinc} models, then finding all valid solutions that satisfy the given constraints. This solver type is well-suited for scheduling and assignment problems, such as matching research papers to conference sessions while respecting topic compatibility and capacity constraints.

\section{Execution Semantics} 
\label{appen:plan_execution}
In this section, we provide a detailed description of the workflow execution process. 
Formally, given the question components $\mathcal{Q}$ and predicted reasoning types $\mathcal{T}$, the {Router} dynamically instantiates and composes an executable workflow $ \pi$. It is part of the Router to assess the required solvers associated with the task, in case there are no suitable solver to solve the task, the Router will instantiate a neural solver modules with LLM. Besides, we define {Memory} as a shared cache for intermediate results among modules during execution. Each module first retrieves the required inputs from {Memory} and write down the running results. By doing this, each module relies solely on {Memory} for its input parameters, which unifies the interface across various modules.

The execution workflow outputs (i) the set of required modules as execution nodes, and (ii) a set of directed edges (source module → target module) as follows:
\begin{equation}
    \pi \;=\;\bigl(V,E,\prec\bigr),
\end{equation}
which is specified by:
\begin{itemize}
  \item $V\subseteq\mathcal{S}\cup{\mathcal{Q}}\cup\widehat{\mathcal{S}}$: a finite set of {execution nodes}.  In this framework, we instantiate each question in $\mathcal{Q}$ as a static module, so as to write the problem data into Memory.
  $\mathcal{S} \;=\;\{\,S_{LP},S_{FOL},S_{CSP},S_{SMT}\}$ denotes the symbolic solver for different reasoning types.
        $\widehat{\mathcal{S}}=\{\hat S_{1},\dots\}$ is an unbounded pool of
        {dynamic solver modules} that may be instantiated on‐the‐fly to solve the problem.
  \item $E\subseteq V\times V$: a set of {directed edges}.  
        $(u,v)\in E$ means “pass the output of $u$ as (part of) the input of $v$”.
  \item $\prec$: a strict total order on $V$ that yields the execution sequence  
        $u_1\prec u_2\prec\cdots\prec u_\ell,\;u_i\in V$.
\end{itemize}

For each module $u \in V$, we assign a unique module id and denote its output as $\boldsymbol{o}(u)$. Let $\textit{in}(v) = \{u \mid (u, v) \in E\}$ denote the set of predecessors of $v$. To ensure that module $v$ receives the correct outputs from its predecessors, each module $u$ stores its output in memory with its agent id, for example, ``\texttt{result\_[ID]}''. Before invoking module $v$, the framework gathers the outputs of all mnodule in $\textit{in}(v)$ with corresponding module ids and assembles them as input to $v$:
\begin{equation}
\boldsymbol{i}(v)
 \;=\;
 \bigl\{\,
   \boldsymbol{o}(u)\,\bigm|\,(u,v)\in\textit{in}(v)
 \bigr\}.    
\end{equation}
The invocation $v(\boldsymbol{i}(v))\to\boldsymbol{o}(v)$ is side-effect free except
for potential updates to $v$’s private \textsc{Memory}.

To better organize the execution results, we introduce two special virtual modules: \texttt{<START>} and \texttt{<END>}. We connect \texttt{<START>} to all source nodes (i.e., nodes with no incoming edges), and direct all terminal nodes (i.e., nodes with no outgoing edges) to \texttt{<END>}. The \texttt{<END>} module aggregates the outputs from all execution paths, ensuring that the overall output is complete. For example, when solving multiple subproblems, the Planner invokes different agents to handle each subproblem, and \texttt{<END>} collects all the predicted results into a list and returns them.

\section{Details of Evaluating Datasets}
\label{append:datasets}

Below we provide the detailed information of the dataset used in this paper.

\begin{enumerate}
   \item \textbf{PrOntoQA} \cite{saparov2023prontoqa} is a synthetic dataset designed to evaluate the deductive reasoning ability of language models. Following the Logic-LM \cite{pan2023logic} setup, we adopt the most challenging subset from PrOntoQA, which involves five-hop reasoning over fictional characters. This subset consists of 500 examples, each requiring the model to determine the truth value of a derived fact.

   \item \textbf{ProofWriter} \cite{tafjord-etal-2021-proofwriter} is a widely used dataset for testing deductive reasoning, where the questions are expressed in natural language. To reduce computational cost, we use the 600-example subset sampled by Logic-LM \cite{pan2023logic} from the depth-5 partition under the open-world assumption. This subset includes balanced labels from {PROVED, DISPROVED, UNKNOWN}, with each instance represented as a (premise set, query) pair.

   \item \textbf{FOLIO} \cite{han-etal-2024-folio} is a challenging dataset for first-order logic reasoning, composed of expert-written problems grounded in real-world knowledge. The examples involve complex semantic structures and require multi-step inference. We evaluate on the full test set, which contains 204 examples.

   \item \textbf{LogDed7} \cite{srivastava2023beyond} is constructed using a subset of the LogicalDeduction dataset from BigBench \cite{srivastava2023beyond}. This task challenges models to determine the sequential ordering of objects based on a minimal set of logical constraints. From the original dataset, we selected the most challenging problems that involve 7 objects, which includes 700 samples.

   \item \textbf{TREC$_{trials}$} \cite{soboroff2021overviewTREC} data including 300 trial-patient pairs, which is built upon the 2021 Clinical Trials Track from TREC to create a trial-patient matching challenge. Our dataset comprises 75 patients, where each patient is paired with both two compatible clinical trials and two incompatible trials. The ground truth matching labels are based on expert annotations and should be considered approximate rather than definitive, since patient descriptions typically do not cover all the variables specified in the trial criteria. As a result, matching decisions often involve informed guesses and therefore depend on experts' personal judgments.

\end{enumerate}
\begin{table*}[h]
\centering
\resizebox{\textwidth}{!}{
\begin{tabular}{ccccc}
\toprule
\textbf{Model} & \textbf{Parameters} & \textbf{Context Length} & \textbf{Key Features/Specialties} & \textbf{License/Access} \\
\midrule
GPT-4o & \makecell{-} & 128K tokens & \makecell{Multimodal, \\ advanced reasoning} & \makecell{API-based \\ (gpt-4o)} \\
\midrule
\makecell{DeepSeek-V3.1} & 671B & 128K tokens & \makecell{Hybrid inference,
 \\ Improved tool / agent capability} &  MIT License \\
\midrule
\makecell{Llama-3.1-8B \\ -Instruct} & 8.03B & 128K tokens & Training: 15T tokens & \makecell{Llama 3.1 \\ Community License} \\
\midrule
\makecell{Qwen-2.5-7B \\ -Instruct} & 7.07B & \makecell{32K tokens \\ (expandable to 128K)} & \makecell{Multilingual, \\ reasoning tasks} & Apache 2.0 \\
\midrule
\makecell{Qwen-2.5-Coder \\ -7B-Instruct} & 7.07B & 32K tokens & \makecell{Code generation, \\ 40+ programming languages} & Apache 2.0 \\
\bottomrule
\end{tabular}
}
\caption{Model Specifications and Characteristics}
\label{tab:model_specs}
\end{table*}

\section{Evaluation and Fine-tuning Details}
\label{append:eval_details}

\subsection{Selection criteria on Foundational models}

\noindent
\textbf{GPT-4o selection:} GPT-4o serves as our primary large-scale baseline due to its state-of-the-art performance across diverse tasks and widespread industry adoption. Its consistent API availability ensures reproducible experimental conditions.

\paragraph{DeepSeek selection:} We selected DeepSeek-V3.1 as the representative large-scale open-source baseline  due to its state-of-the-art performance for reasoning. 

\paragraph{7B-8B parameter range:} We selected models in the 7B-8B parameter range as they represent the optimal balance between computational feasibility and performance capability. This size range enables deployment on single high-end GPUs while maintaining sophisticated reasoning abilities, making them practical alternatives to larger models.

\paragraph{Model diversity:} Our selection encompasses proprietary (GPT-4o) versus open-source models, general-purpose versus code-specialized variants, and different institutional approaches (DeepSeek-AI Team, OpenAI Team, Meta Team, and Qwen Team). The 7B-8B range enables direct comparison between open-source alternatives (Llama-3.1-8B vs. Qwen-2.5-7B) and evaluation of specialization effects (Qwen-2.5-7B vs. Qwen-2.5-Coder-7B).

Table \ref{tab:model_specs} presents the specifications and characteristics of each model.

\subsection{Evaluation Details}
Here we provide the LLM settings compared in this paper.

\begin{enumerate}
   \item \textbf{Zero-shot}: In the zero-shot setting, the LLM is asked to directly provide the final answer. 
   \item \textbf{Chain-of-Thought} (CoT): In the CoT setting, the LLM is instructed to output both the answer and a step-by-step reasoning process.
   \item \textbf{Symbolic CoT} (CoT$_{sym}$): In the Symbolic CoT setting, the LLM is first prompted to translate the problem into a formal language representation, and then perform reasoning with LLM based on the formalization.
\end{enumerate}

Additionally, to validate the effectiveness of the routing mechanism within our framework, we integrate the Router's outputs as auxiliary prompts into the aforementioned baseline settings.
Specifically, we prepend the Router’s output to the input prompt as an auxiliary hint before the original question, allowing the LLM to perform reasoning with this additional guidance. In particular, for CoT$_{sym}$, after the Router invokes the corresponding solver, we directly provide the resulting formalization to the LLM, instead of requiring the model to generate it on its own. This setup enables us to observe how the \texttt{Router} influences model performance across different prompting strategies.

For \texttt{GPT-4o}, we use the default API parameters to obtain responses. For open-source models, we set the maximum number of new tokens to 4096 and the temperature to 0.01. Besides, all models are required to return their predictions in JSON format using the keyword ``\texttt{ans}". For the CoT and Symbolic CoT settings, models are also required to include their reasoning process with the keyword ``\texttt{reasoning}". 

To manage inference costs, we evaluate {GPT-4o} and {DeepSeek-V3.1} using pass@1 metric, which measures accuracy based on a single inference run.For each small model, since different settings use the same pretrained version for both routing and parsing, we perform routing and parsing once, and share the results across all inference settings. We conduct experiments three times with different random seeds and report the average performance across the three runs.

\begin{table}[!t]
\centering
\resizebox{\linewidth}{!}{
\begin{tabular}{ccc}
\toprule
\textbf{Dataset} & \textbf{Solver Type} & \textbf{\# Examples} \\
\midrule
ProntoQA & LP Solver & 1,390 \\
ProofWriter & LP Solver & 1,398 \\
FOLIO & FOL Solver & 776 \\
Logical Deduction & CSP Solver & 1,185 \\
TREC & SMT Solver & 376 \\
\midrule
\textbf{Total} & \textbf{All Solvers} & \textbf{5,125} \\
\bottomrule
\end{tabular}
}
\caption{Training Data Statistics for Each Reasoning Engine}
\label{tab:training_data}
\end{table}

\subsection{Fine-tuning details}
Due to the poor understanding capabilities of open-source models for non-mainstream programming languages, we constructed formalization datasets for relevant reasoning engines and fine-tuned these models to enhance their performance within the our framework framework.

Our training data is primarily constructed according to the tasks used in our experiments. We begin by selecting problems for formalization across different reasoning engines:
\begin{itemize}
    \item For the \textbf{LP solver}, we use problems from the training portion of ProofWriter provided by Logic-LM and one-hop problems from ProntoQA.
    \item For the \textbf{FOL solver}, we select the training portion of FOLIO provided by Logic-LM.
    \item For the \textbf{CSP solver}, we use the training portion of LogDed provided by Logic-LM.
    \item For the \textbf{SMT solver}, we use the TREC 2022 Clinical Trials Track.
\end{itemize}

For each problem, we utilize GPT-4o to obtain the corresponding formalization code for each reasoning engine. To enhance data quality, we execute the formalization with the corresponding reasoning engines and retain only the code that runs successfully. The final data quantities obtained for each component are shown in Table \ref{tab:training_data}.

Using the Llamafactory \cite{zheng2024llamafactory} training framework, we employed LoRA fine-tuning for all models to prevent overfitting given the limited training samples. The LoRA configuration was set with $\text{rank}=16$ and applied to all modules. 
For all models, we set the \texttt{sequence\_length} to $3072$, \texttt{per\_device\_train\_batch\_size} to $2$, and \texttt{gradient\_accumulation\_steps} to $8$. 
The \texttt{learning\_rate} was selected from the set $\{1e-3, 1e-4\}$ based on optimal performance on the validation set. All models were trained for three epochs.  All training experiments are conducted on two H100 GPUs.

\begin{table*}[!ht]
  \small
    \centering
    \setlength {\tabcolsep }{1mm}
    \resizebox{\textwidth}{!}{
      \begin{tabular}{c|cc|cc|cc|cc|cc|cc|c}
      \toprule
      \multirow{2}[0]{*}{\textbf{Methods}} & \multicolumn{2}{c|}{\textbf{PrOntoQA}} & \multicolumn{2}{c|}{\textbf{ProofWriter}} & \multicolumn{2}{c|}{\textbf{FOLIO}} & \multicolumn{2}{c|}{\textbf{LogDed7}} & \multicolumn{2}{c|}{\textbf{TREC}$_{trials}$} & \multicolumn{2}{c|}{\textbf{Mixed}} & \multirow{2}[0]{*}{\textbf{Routing}} \\
            & w/o routing & w/ routing & w/o routing & w/ routing & w/o routing & w/ routing & w/o routing & w/ routing & w/o routing & w/ routing & w/o routing & w/ routing &  \\
      \midrule
    \multicolumn{14}{c}{\textbf{Llama-3.1-8b}} \\
\midrule
Zero-shot & 65.2$\pm$0.5 & 58.6$\pm$0.3 & 41.8$\pm$0.2 & 36.3$\pm$0.1 & 46.7$\pm$0.2 & 42.6$\pm$0.8 & 37.0$\pm$0.1 & 25.3$\pm$0.2 & 63.2$\pm$0.2 & 39.7$\pm$0.9 & 48.7$\pm$0.1 & 38.8$\pm$0.3 & \multirow{5}{*}{76.8$\pm$2.7} \\
CoT & 57.9$\pm$0.7 & 54.1$\pm$0.7 & 45.3$\pm$0.8 & 35.4$\pm$1.2 & 38.2$\pm$0.4 & 42.3$\pm$0.2 & 19.5$\pm$0.6 & 24.0$\pm$1.4 & 63.6$\pm$0.6 & 48.3$\pm$0.5 & 42.0$\pm$0.5 & 38.3$\pm$0.5 &   \\
CoT$_{sym}$ & 10.1$\pm$1.0 & 0.0$\pm$0.0 & 10.8$\pm$0.6 & 0.2$\pm$0.2 & 12.1$\pm$1.7 & 0.2$\pm$0.2 & 3.9$\pm$0.2 & 0.6$\pm$0.1 & 2.6$\pm$0.4 & 3.1$\pm$0.4 & 7.6$\pm$0.3 & 0.7$\pm$0.1 &   \\
Ours & - & 3.9$\pm$1.4 & - & 5.3$\pm$1.1 & - & 0.8$\pm$0.5 & - & 7.0$\pm$0.2 & - & 27.6$\pm$1.7 & - & 8.0$\pm$0.9 &   \\
Ours$_{sft}$ & - & 75.7$\pm$13.9 & - & 51.5$\pm$11.8 & - & 31.5$\pm$5.0 & - & 10.1$\pm$1.4 & - & 40.9$\pm$2.8 & - & 41.0$\pm$7.3 &   \\
    \midrule
    \multicolumn{14}{c}{\textbf{Qwen-2.5-7b}} \\
\midrule
Zero-shot & 57.2$\pm$0.0 & 40.9$\pm$0.2 & 37.8$\pm$0.1 & 29.8$\pm$0.2 & 32.4$\pm$0.0 & 40.0$\pm$0.2 & 42.9$\pm$0.1 & 38.5$\pm$0.4 & 59.3$\pm$0.0 & 53.2$\pm$0.2 & 45.9$\pm$0.0 & 38.8$\pm$0.2 & \multirow{5}{*}{97.7$\pm$0.1} \\
CoT & 95.3$\pm$0.4 & 72.8$\pm$0.2 & 59.8$\pm$0.7 & 59.1$\pm$0.1 & 48.0$\pm$1.2 & 47.7$\pm$1.7 & 46.8$\pm$1.2 & 20.6$\pm$0.7 & 47.0$\pm$0.8 & 25.3$\pm$0.3 & 60.8$\pm$0.3 & 45.0$\pm$0.1 &   \\
CoT$_{sym}$ & 55.1$\pm$0.1 & 46.1$\pm$0.2 & 41.7$\pm$0.2 & 29.7$\pm$0.6 & 59.2$\pm$0.2 & 52.3$\pm$0.5 & 8.0$\pm$0.6 & 49.5$\pm$0.8 & 60.2$\pm$0.3 & 61.9$\pm$1.1 & 38.3$\pm$0.2 & 45.5$\pm$0.2 &   \\
Ours & - & 0.0$\pm$0.0 & - & 0.0$\pm$0.0 & - & 0.0$\pm$0.0 & - & 31.5$\pm$0.3 & - & 23.1$\pm$0.6 & - & 12.6$\pm$0.2 &   \\
Ours$_{sft}$ & - & 71.8$\pm$8.2 & - & 51.9$\pm$5.6 & - & 6.4$\pm$1.2 & - & 45.0$\pm$0.4 & - & 51.9$\pm$2.2 & - & 50.1$\pm$3.2 &   \\
    \midrule
    \multicolumn{14}{c}{\textbf{Qwen-2.5-Coder-7b}} \\
    \midrule
Zero-shot & 71.5$\pm$0.1 & 67.8$\pm$0.0 & 53.6$\pm$0.2 & 56.5$\pm$0.2 & 45.3$\pm$0.2 & 25.7$\pm$0.5 & 45.8$\pm$0.2 & 44.1$\pm$0.1 & 66.2$\pm$0.2 & 66.7$\pm$0.3 & 56.0$\pm$0.1 & 53.8$\pm$0.1 & \multirow{5}{*}{92.9$\pm$0.1} \\
CoT & 70.5$\pm$0.6 & 64.8$\pm$0.3 & 51.3$\pm$0.1 & 53.3$\pm$0.3 & 45.4$\pm$0.2 & 24.8$\pm$0.6 & 40.5$\pm$0.4 & 40.4$\pm$0.3 & 63.0$\pm$0.3 & 65.2$\pm$0.6 & 53.2$\pm$0.1 & 50.9$\pm$0.2 &   \\
CoT$_{sym}$ & 78.2$\pm$14.6 & 29.8$\pm$0.3 & 46.8$\pm$9.6 & 20.6$\pm$1.0 & 52.1$\pm$9.5 & 13.2$\pm$0.7 & 14.0$\pm$2.7 & 11.3$\pm$1.0 & 38.0$\pm$8.3 & 29.3$\pm$2.6 & 43.0$\pm$8.4 & 20.3$\pm$0.3 &   \\
Ours & - & 6.1$\pm$0.2 & - & 12.8$\pm$0.5 & - & 2.8$\pm$0.2 & - & 60.2$\pm$0.4 & - & 23.2$\pm$1.0 & - & 26.2$\pm$0.1 &   \\
Ours$_{sft}$ & - & 89.4$\pm$0.2 & - & 44.9$\pm$0.5 & - & 3.1$\pm$0.5 & - & 74.2$\pm$0.2 & - & 43.3$\pm$1.2 & - & 59.6$\pm$0.1 &   \\
    \bottomrule
  \end{tabular}%
  }
  \caption{Average performance of small open-source models over three random seeds with standard deviations.}
  \label{tab:empirical_evaluation_small_std}
\end{table*}%

\section{Complete Results for Small Models Evaluation}
\label{appen:complete_res}
For small-scale model evaluation in Section \ref{sec:res_on_small}, we conduct experiments with 3 random seeds and report averaged performance. Here we present the complete results, including means and standard deviations in Table \ref{tab:empirical_evaluation_small_std}.

\section{Prompt for Text Parsing}
\label{appen:prompt_parser}

We use the prompt in Figure \ref{fig:appen_prompt_parsing_1} and Figure \ref{fig:appen_prompt_parsing_2} to decompose the input text.

\begin{figure*}[htbp]
\centering
\tcbset{
  colback=white,
  colframe=black,
  boxrule=0.5pt,
  arc=0mm,
  width=\textwidth,
}

\begin{tcolorbox}[title={Prompt for Text Parsing: Part 1}]
  \noindent
  \ttfamily
  \footnotesize
  \begin{minipage}[t]{\textwidth}
SYSTEM:
You are a logician and reasoning systems expert specializing in symbolic reasoning frameworks. Given a text that may contain one or multiple logical reasoning problems, identify each problem, determine its type, and decompose the text accordingly. Return the result strictly as a JSON object with "result" containing an array of problem objects.

Specifically, your task is to:

1. First, analyze the input text to identify how many distinct reasoning problems it contains.

2. For each identified problem, determine its type from the following categories:
    
    - Logic Programming (LP): Problems where conclusions are typically deduced step by step from a set of known facts and rules. These problems often involve applying simple logical rules repeatedly to infer new facts until the goal is reached.
    
    - First-order Logic (FOL): Problems that require more expressive reasoning, such as statements like "for all" or "there exists", and complex relationships among multiple entities. 
    
    - Constraint Satisfaction Problem (CSP): Problems that involve finding assignments of values to variables within finite domains such that all explicit or implicit constraints are satisfied. These often include tasks like ordering, allocation, or scheduling.
   
    - Boolean Satisfiability (SAT): These problems involve determining whether all logical constraints in a system are simultaneously satisfied. In the context of reasoning tasks, SAT typically focuses on checking if a particular configuration or entity conforms to a set of conditions expressed as logical formulas. Unlike CSP, which searches for value assignments over finite domains, SAT emphasizes verifying logical consistency in given assignments and is often applied to analytical reasoning questions.

To guide your classification:
    
    - Consider whether the problem leans more towards reasoning from facts and rules (often LP or FOL) or checking constraints and conditions (often CSP or SAT).
    
    - If the focus is on assigning values or arranging elements under constraints, it is more typical of CSP.
    
    - If the focus is on verifying whether one given description satisfies the logical requirements of another, it is more typical of SAT. Analytical reasoning problems are often classified as SAT.
    
    - Between LP and FOL, use LP if the reasoning relies on simple rules and chaining; use FOL if it involves richer logical expressions with quantifiers or complex entity relationships.

3. For each problem, create a JSON object with the following structure:

- "problem\_id" (str): A unique identifier following the pattern "ques\_1", "ques\_2", etc., based on the order of appearance.

- "problem\_type" (str): The type classification. The value must be one of {LP, FOL, CSP, SAT}.

- Based on the problem type, include the appropriate fields:
  
  - If problem\_type == "LP" or "FOL":
  
    - "premise" (str): the given premise.
    
    - "hypothesis" (str): the hypothesis to be evaluated for truth.
    
    - "options" (list): the provided answer options.
  
  - If problem\_type == "CSP":
    
    - "context" (str): background description.
    
    - "question" (str): the specific question being asked.
    
    - "options" (list): the provided answer options.
  
  - If problem\_type == "SAT":
    
    - "trial\_description" (str): description of the trial.
    
    - "sample\_description" (str): description of the sample.
    
    - "options" (list): the provided answer options.

Preserve any existing option labels (e.g., "A)", "B)"). If options have no labels, assign labels {'A)', 'B)', 'C)', ...} automatically.

(Part 2 of the prompt will follow.)
  \end{minipage}
\end{tcolorbox}
\caption{Prompt for text parsing.}
\label{fig:appen_prompt_parsing_1}
\end{figure*}

\begin{figure*}[htbp]
\centering
\tcbset{
  colback=white,
  colframe=black,
  boxrule=0.5pt,
  arc=0mm,
  width=\textwidth,
}

\begin{tcolorbox}[title={Prompt for Text Parsing: Part 2}]
  \noindent
  \ttfamily
  \footnotesize
  \begin{minipage}[t]{\textwidth}
(Following Part 1 of the prompt for text parsing.)

4. Extract or analyze the overall goal of the input text:

- FIRST, try to extract any explicitly stated overall goal or instruction from the text (e.g., "Answer the above questions one by one", "Solve all problems to find the final answer", etc.)

- If no explicit goal is found, analyze the relationship between problems and write a brief description:
  
  - Multiple independent problems: "Solve multiple independent reasoning problems"
  
  - Subproblems contributing to main problem: "Solve subproblems to address the main complex problem" 
  
  - Sequential dependent problems: "Solve problems in sequence with dependencies"

  - Single problem: "Solve the reasoning problem"

Return a JSON object with two keys:

- "result": an array containing all identified problems

- "overall\_goal": the extracted goal text or a brief analysis-based description

Example output format:

{

  "result": [
    
    {
      
      "problem\_id": "ques\_1",
      
      "problem\_type": "CSP",
      
      "context": "...",
      
      "question": "...", 
    
      "options": ["A) ...", "B) ..."]
    
    }
    
  ],

  "overall\_goal": "Answer the above questions one by one"

}

USER: 

Problem Statement:

{problem}
  \end{minipage}
\end{tcolorbox}
\caption{Prompt for text parsing.}
\label{fig:appen_prompt_parsing_2}
\end{figure*}

\section{Prompt for Routing}

The \texttt{Router} uses prompt in Figure \ref{fig:appen_planning} to select task-relevant agents and construct an execution plan.

\begin{figure*}[htbp]
\centering
\tcbset{
  colback=white,
  colframe=black,
  boxrule=0.5pt,
  arc=0mm,
  width=\textwidth,
}

\begin{tcolorbox}[title={Prompt for Routing}]
  \noindent
  \ttfamily
  \small
  \begin{minipage}[t]{\textwidth}
Design a plan that uses the minimal number of agents necessary to achieve the goals. You may be given one or multiple problems to solve, each with a unique problem\_id. Select agents only from the provided list. Output a JSON object describing the plan.Requirements:

1. Use the MINIMAL number of agents needed to complete all tasks.

2. The output MUST be a JSON object with exactly two keys:
  
  • 'agents': an array of selected agent names (strings), including any problem\_ids that serve as starting points

  • 'edges': an array of [source, target] pairs (both strings) showing execution order.

3. For multi-problem scenarios, start the execution flow from the respective problem\_ids.

4. The plan MUST end with the special control marker <END>.

5. IMPORTANT: If you need to use the same agent type for multiple different problems, distinguish them by adding ':' followed by a sequence number. For example, if you need two CSP solvers for different problems, use 'csp\_solver:1' and 'csp\_solver:2'.

6. You can use the same agent for multiple problems if appropriate, but ensure proper sequencing.

Example for multiple problems:

If given 'QID [ques\_1]: CSP scheduling problem', 'QID [ques\_2]: FOL reasoning task' and 'QID [ques\_3]: CSP allocation problem', your output might include agents: ['ques\_1', 'ques\_2', 'ques\_3', 'csp\_solver:1', 'fol\_solver:1', 'csp\_solver:2', '<END>'] with edges: [['ques\_1', 'csp\_solver:1'], ['ques\_2', 'fol\_solver:1'], ['ques\_3', 'csp\_solver:2'], ['csp\_solver:1', '<END>'], ['fol\_solver:1', '<END>'], ['csp\_solver:2', '<END>']].

7. Respond with **only valid JSON**, without any explanations, markdown formatting, or code fences. Do not wrap the output in ```json or any other delimiters. Return pure JSON.

Here is the goal: [GOAL]

Problems to solve: [PROBLEMS]

Current agents: 

[PORTFOLIO]

  \end{minipage}
\end{tcolorbox}
\caption{Prompt for routing.}
\label{fig:appen_planning}
\end{figure*}

\section{Prompt Templates for Textual Instance Construction}

\label{appen:prompt_for_merge}

To support end-to-end evaluation where models does not have the prior knowledge of task-specific information, we convert each data instance into the natural language format using dataset-specific prompt templates. These templates integrate the key components of each problem (e.g., premises, hypotheses, questions) into a single input text.

Specifically, for PrOntoQA, ProofWriter, and FOLIO, we apply the following templates to convert structured fields into natural language text.

{
\centering
\tcbset{
  colback=white,
  colframe=black,
  boxrule=0.5pt,
  arc=0mm,
  width=\linewidth,
}

\begin{tcolorbox}[title={Prompt}]
  \noindent
  \ttfamily
  \small
  \begin{minipage}[t]{\textwidth}
STATEMENT: \\
\{premise\} \\

QUESTION: \\
\{conclusion\} \\

\{options\}
  \end{minipage}
\end{tcolorbox}
}

For CSP task, we use the following template:

{
\centering
\tcbset{
  colback=white,
  colframe=black,
  boxrule=0.5pt,
  arc=0mm,
  width=\linewidth,
}

\begin{tcolorbox}[title={Prompt}]
  \noindent
  \ttfamily
  \small
  \begin{minipage}[t]{\textwidth}
STATEMENT: \\
\{input\} \\

Which of the following is true? \\
\{options\}
  \end{minipage}
\end{tcolorbox}
}

For SMT task, we use the following template:

{
\centering
\tcbset{
  colback=white,
  colframe=black,
  boxrule=0.5pt,
  arc=0mm,
  width=\linewidth,
}

\begin{tcolorbox}[title={Prompt}]
  \noindent
  \ttfamily
  \small
  \begin{minipage}[t]{\textwidth}
You get a trial and a patient and have to say if there is a match: \\

TRIAL: \{trial\_description\} \\

PATIENT: \{patient\_description\} \\

Does the patient match the trial? \\
A) True \\
B) False
  \end{minipage}
\end{tcolorbox}
}

Each prompt is filled with instance-specific content to form the complete input, which is then passed to the model without revealing dataset boundaries or task types. This ensures the evaluation closely reflects practical use cases where models must perform problem-type identification and reasoning jointly.

\section{Prompt Templates for Generating Multi-Question Prompts}
\label{appen:sec_prompt_template}
We use the prompt format in Figure \ref{fig:appen_merge_ques} to concatenate multiple questions into a single input prompt.

\begin{figure}[!t]
\centering
\tcbset{
  colback=white,
  colframe=black,
  boxrule=0.5pt,
  arc=0mm,
  width=\linewidth,
}

\begin{tcolorbox}[title={Prompt for Merging Questions}]
  \noindent
  \ttfamily
  \small
  \begin{minipage}[t]{\textwidth}
Answer the following questions one by one.

Q1:\{question1\}

Q2:\{question2\}

Q3:\{question3\}

  \end{minipage}
\end{tcolorbox}
\caption{Prompt for merging questions.}
\label{fig:appen_merge_ques}
\end{figure}

\section{Example for Clinical Trial Matching}

An example of an obtained z3 formalization of clinical trial-patient pair.

\begin{figure}[!t]
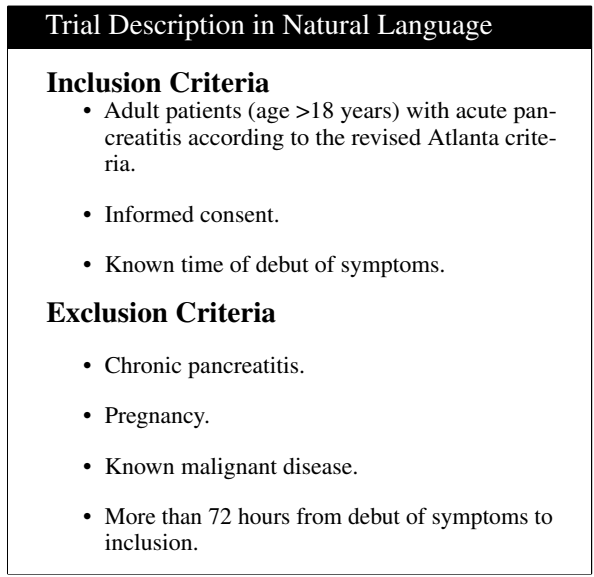

\centering
\tcbset{
  colback=white,
  colframe=black,
  boxrule=0.5pt,
  arc=0mm,
  width=\linewidth,
}

\begin{tcolorbox}[title={Trial Description in Natural Language}]
  \noindent
  \small
  \begin{minipage}[htbp]{\textwidth}
\paragraph{Inclusion Criteria}
\begin{itemize}
  \item Adult patients (age >18 years) with acute pancreatitis according to the revised Atlanta criteria.
  \item Informed consent.
  \item Known time of debut of symptoms.
\end{itemize}

\paragraph{Exclusion Criteria}
\begin{itemize}
  \item Chronic pancreatitis.
  \item Pregnancy.
  \item Known malignant disease.
  \item More than 72 hours from debut of symptoms to inclusion.
\end{itemize}
  \end{minipage}
\end{tcolorbox}
\caption{Trial Description in natural language.}
\label{fig:appen_trial_desc_nl}
\end{figure}

\begin{figure}[!t]
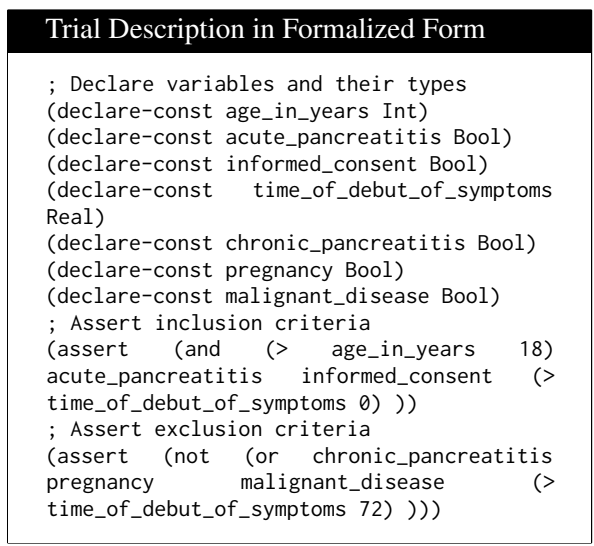

\centering
\tcbset{
  colback=white,
  colframe=black,
  boxrule=0.5pt,
  arc=0mm,
  width=\linewidth,
}

\begin{tcolorbox}[title={Trial Description in Formalized Form}]
  \noindent
  \ttfamily
  \small
  \begin{minipage}[htbp]{\textwidth}
; Declare variables and their types

(declare-const age\_in\_years Int)

(declare-const acute\_pancreatitis Bool)

(declare-const informed\_consent Bool)

(declare-const time\_of\_debut\_of\_symptoms Real)

(declare-const chronic\_pancreatitis Bool)

(declare-const pregnancy Bool)

(declare-const malignant\_disease Bool)

; Assert inclusion criteria

(assert (and
  (> age\_in\_years 18)
  acute\_pancreatitis
  informed\_consent
  (> time\_of\_debut\_of\_symptoms 0)
))

; Assert exclusion criteria

(assert (not (or
  chronic\_pancreatitis
  pregnancy
  malignant\_disease
  (> time\_of\_debut\_of\_symptoms 72)
)))
  \end{minipage}
\end{tcolorbox}
\caption{Trial Description in formalized form.}
\label{fig:appen_trial_desc_form}
\end{figure}

As shown in Figure \ref{fig:appen_trial_desc_nl} and Figure \ref{fig:appen_trial_desc_form}, formalizing the clinical trial description involves declaring all mentioned variables and the eligibility rules—which are often divided into inclusion criteria and exclusion criteria.

\begin{figure}[!t]
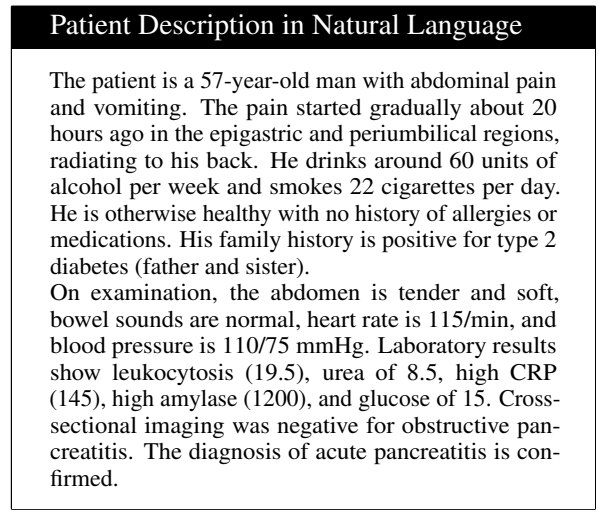

\centering
\tcbset{
  colback=white,
  colframe=black,
  boxrule=0.5pt,
  arc=0mm,
  width=\linewidth,
}

\begin{tcolorbox}[title={Patient Description in Natural Language}]
  \noindent
  \small
  \begin{minipage}[htbp]{\textwidth}
The patient is a 57-year-old man with abdominal pain and vomiting. The pain started gradually about 20 hours ago in the epigastric and periumbilical regions, radiating to his back. He drinks around 60 units of alcohol per week and smokes 22 cigarettes per day. He is otherwise healthy with no history of allergies or medications. His family history is positive for type 2 diabetes (father and sister). 

On examination, the abdomen is tender and soft, bowel sounds are normal, heart rate is 115/min, and blood pressure is 110/75 mmHg. Laboratory results show leukocytosis (19.5), urea of 8.5, high CRP (145), high amylase (1200), and glucose of 15. Cross-sectional imaging was negative for obstructive pancreatitis. The diagnosis of acute pancreatitis is confirmed.
  \end{minipage}
\end{tcolorbox}
\caption{Patient description in natural language.}
\label{fig:appen_patient_desc_nl}
\end{figure}

\begin{figure}[!t]
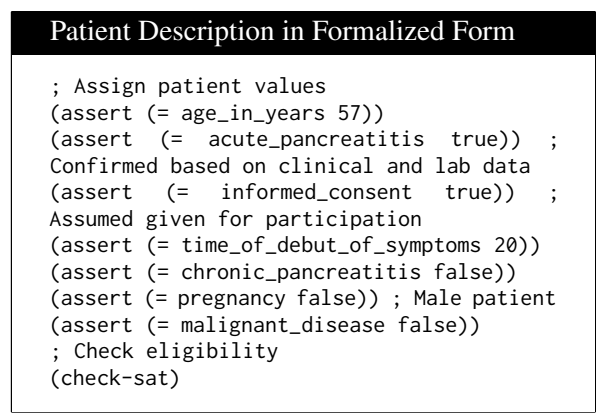

\centering
\tcbset{
  colback=white,
  colframe=black,
  boxrule=0.5pt,
  arc=0mm,
  width=\linewidth,
}

\begin{tcolorbox}[title={Patient Description in Formalized Form}]
  \noindent
  \small
  \ttfamily
  \begin{minipage}[htbp]{\textwidth}
; Assign patient values

(assert (= age\_in\_years 57))

(assert (= acute\_pancreatitis true))   ; Confirmed based on clinical and lab data

(assert (= informed\_consent true))     ; Assumed given for participation

(assert (= time\_of\_debut\_of\_symptoms 20))

(assert (= chronic\_pancreatitis false))

(assert (= pregnancy false))            ; Male patient

(assert (= malignant\_disease false))

; Check eligibility

(check-sat)
  \end{minipage}
\end{tcolorbox}
\caption{Patient in formalized form.}
\label{fig:appen_patient_desc_form}
\end{figure}

The patient description needs to be formalized accordingly: in such a way that the z3 code can run on the trial-patient pair, as shown in Figure \ref{fig:appen_patient_desc_nl} and \ref{fig:appen_patient_desc_form}.

Note that the SMT only sets the parameter values that are present in the formalized clinical trials description, as any other values will not play any role in determining if there is a match.

The agent also adds comments to justify the parameter values it sets. This is important, as it will often need to infer likely values that are not mentioned in the patient description. Indeed, in our example, the trial requires a confirmed diagnosis of acute pancreatitis according to the revised Atlanta criteria, but the patient description only provides indirect evidence—such as elevated amylase, characteristic pain, and inflammatory markers. The model therefore infers that the patient meets the diagnostic criteria and sets the variable for acute pancreatitis to true. Note that such cases of unmentioned parameters (that thus require abductive inferences) are the norm in the TREC dataset—and one of the reason the truth matching labels are approximate.

\section{Example for Other Solvers}
Besides the SMT solver, here we also present formalization examples for the remaining solvers. Figure \ref{fig:appen_lp_solver}, Figure \ref{fig:appen_fol_solver}, and Figure \ref{fig:appen_csp_solver} demonstrate the formalization processes for the LP solver, FOL solver, and CSP solver, respectively.

\begin{figure*}[htbp]
\centering
\tcbset{
  colback=white,
  colframe=black,
  boxrule=0.5pt,
  arc=0mm,
  width=\textwidth,
}

\begin{tcolorbox}[title={Formalization of LP Solver}]
  \noindent
  \small
  \begin{minipage}[t]{\textwidth}
\textbf{Premise:}\\
Every dumpus is not red. Tumpuses are red. Dumpuses are impuses. Impuses are not feisty. Impuses are yumpuses. Stella is a dumpus.\\
\\
\textbf{Hypothesis:}\\
Stella is not red.\\
\\
\textbf{Formalization:}

{
  \ttfamily
Predicates:\\
dumpus(\$x, bool) ::: Is x a dumpus?\\
red(\$x, bool) ::: Is x red?\\
tumpus(\$x, bool) ::: Is x a tumpus?\\
impus(\$x, bool) ::: Is x an impus?\\
feisty(\$x, bool) ::: Is x feisty?\\
yumpus(\$x, bool) ::: Is x a yumpus?\\
isA(\$x, \$type, bool) ::: Is x of type \$type?\\
\\
Facts:\\
dumpus(Stella, True) ::: Stella is a dumpus.\\
\\
Rules:\\
dumpus(\$x, True) >>> red(\$x, False) ::: Every dumpus is not red.\\
tumpus(\$x, True) >>> red(\$x, True) ::: Tumpuses are red.\\
dumpus(\$x, True) >>> impus(\$x, True) ::: Dumpuses are impuses.\\
impus(\$x, True) >>> feisty(\$x, False) ::: Impuses are not feisty.\\
impus(\$x, True) >>> yumpus(\$x, True) ::: Impuses are yumpuses.\\
\\
Query:\\
red(Stella, False) ::: Stella is not red.
}

  \end{minipage}
\end{tcolorbox}
\caption{Formalization of LP Solver.}
\label{fig:appen_lp_solver}
\end{figure*}

\begin{figure*}[htbp]
\centering
\tcbset{
  colback=white,
  colframe=black,
  boxrule=0.5pt,
  arc=0mm,
  width=\textwidth,
}

\begin{tcolorbox}[title={Formalization of FOL Solver}]
  \noindent
  \small
  \begin{minipage}[t]{\textwidth}
\textbf{Premise:}\\
All birds have wings. No animal with wings is a reptile. Some animals that fly are birds. If something is an iguana, then it is a reptile. John is neither an iguana nor a bird.\\
\\
\textbf{Hypothesis:}\\
John is a reptile.\\
\\
\textbf{Formalization:}

{
  \ttfamily
Predicates:\\
Bird(x) ::: x is a bird.\\
Wings(x) ::: x has wings.\\
Animal(x) ::: x is an animal.\\
Reptile(x) ::: x is a reptile.\\
Fly(x) ::: x can fly.\\
Iguana(x) ::: x is an iguana.\\
\\
Premises:\\
$\forall$x (Bird(x) → Wings(x)) ::: All birds have wings.\\
$\forall$x (Wings(x) → ¬Reptile(x)) ::: No animal with wings is a reptile.\\
$\exists$x (Fly(x) $\wedge$ Bird(x)) ::: Some animals that fly are birds.\\
$\forall$x (Iguana(x) → Reptile(x)) ::: If something is an iguana, then it is a reptile.\\
¬Iguana(john) $\wedge$ ¬Bird(john) ::: John is neither an iguana nor a bird.\\
\\
Conclusions:\\
Reptile(john) ::: John is a reptile.
}

  \end{minipage}
\end{tcolorbox}
\caption{Formalization of FOL Solver.}
\label{fig:appen_fol_solver}
\end{figure*}

\begin{figure*}[htbp]
\centering
\tcbset{
  colback=white,
  colframe=black,
  boxrule=0.5pt,
  arc=0mm,
  width=\textwidth,
}

\begin{tcolorbox}[title={Formalization of CSP Solver}]
  \noindent
  \small
  \begin{minipage}[t]{\textwidth}
\textbf{Premise:}\\
The following paragraphs each describe a set of five objects arranged in a fixed order. The statements are logically consistent within each paragraph.\\\\
In a golf tournament, there were five golfers: Rob, Ada, Dan, Joe, and Mel. Ada finished above Mel. Mel finished above Dan. Joe finished below Dan. Ada finished second.\\
\\
\textbf{Formalization:}

{
  \ttfamily
Formalization:\\
include "globals.mzn";\\
\\
\% Define the golfers\\
enum GOLFER = {Rob, Ada, Dan, Joe, Mel};\\
\\
\% Each golfer has a unique finishing position from 1 (first) to 5 (last)\\
array[GOLFER] of var 1..5: pos;\\
\\
\% All golfers must finish in distinct positions\\
constraint all\_different([pos[g] | g in GOLFER]);\\
\\
\% Ada finished above Mel\\
constraint pos[Ada] < pos[Mel];\\
\\
\% Mel finished above Dan\\
constraint pos[Mel] < pos[Dan];\\
\\
\% Joe finished below Dan\\
constraint pos[Joe] > pos[Dan];\\
\\
\% Ada finished second\\
constraint pos[Ada] = 2;\\
\\
solve satisfy;\\
}

\textbf{Ordering}\\
left to right: (1 = leftmost, 5 = rightmost)

  \end{minipage}
\end{tcolorbox}
\caption{Formalization of CSP Solver.}
\label{fig:appen_csp_solver}
\end{figure*}

\end{document}